\documentclass[10pt,twocolumn,letterpaper]{article}

\usepackage[pagenumbers]{iccv} % To force page numbers, e.g. for an arXiv version

\usepackage{multirow}
\usepackage[normalem]{ulem}

\definecolor{iccvblue}{rgb}{0.21,0.49,0.74}
\usepackage[pagebackref,breaklinks,colorlinks,allcolors=iccvblue]{hyperref}

\def\paperID{14874} % *** Enter the Paper ID here
\def\confName{ICCV}
\def\confYear{2025}

\title{Incremental Human-Object Interaction Detection with\\ Invariant Relation Representation Learning}

\author{Yana Wei$^{1,}$\thanks{Both authors contributed equally to this work.}\quad Zeen Chi$^{1,}$\footnotemark[1]\quad Chongyu Wang$^{1}$ \ Yu Wu$^{1}$ \ Shipeng Yan$^{1}$ \ Yongfei Liu$^{1}$ \ Xuming He$^{1,2}$\\
$^1$ShanghaiTech University\ $^2$Shanghai Engineering Research Center of Intelligent Vision and Imaging\\
{\tt\small\{wynnaggy,zeenchi.2002\}@gmail.com\quad \{wangchy12024,hexm\}@shanghaitech.edu.cn}
}

\begin{document}
\maketitle
\begin{abstract}

    In open-world environments, human-object interactions (HOIs) evolve continuously, challenging conventional closed-world HOI detection models. Inspired by humans' ability to progressively acquire knowledge, we explore incremental HOI detection (IHOID) to develop agents capable of discerning human-object relations in such dynamic environments. 
    This setup confronts not only the common issue of catastrophic forgetting in incremental learning but also distinct challenges posed by interaction drift and detecting zero-shot HOI combinations with sequentially arriving data. 
    Therefore, we propose a novel exemplar-free incremental relation distillation (IRD) framework. IRD decouples the learning of objects and relations, and introduces two unique distillation losses for learning invariant relation features across different HOI combinations that share the same relation. Extensive experiments on HICO-DET and V-COCO datasets demonstrate the superiority of our method over state-of-the-art baselines in mitigating forgetting, strengthening robustness against interaction drift, and generalization on zero-shot HOIs. Code is available at \texttt{\url{https://github.com/weiyana/ContinualHOI}}.

    % \textcolor{red}{Human-object interaction (HOI) detection requires AI agents to analyze how humans interact with objects. Understanding the dynamic world demands continuous learning of such structured information. While incremental learning has been explored in classification and detection, HOI perception remains underexplored in ever-changing environments. We introduce the incremental HOI detection (\textbf{IHOID}) setup, where models must continuously learn new HOIs while retaining past knowledge.
    % This task faces catastrophic forgetting, interaction drift, and zero-shot HOI generalization with sequentially arriving data. We propose an exemplar-free Incremental Relation Distillation (\textbf{IRD}) framework to address these challenges. IRD disentangles object and relation learning and introduces two unique distillation losses for learning invariant relation features across different HOI combinations sharing the same relation.
    % Experiments on HICO-DET and V-COCO show that IRD significantly outperforms state-of-the-art baselines, reducing forgetting, enhancing model robustness to changing data distributions and unseen HOIs.}

\end{abstract}

\section{Introduction}
\label{sec:intro}
% propose the task setup
% The task of human-object interaction (HOI) detection \cite{zhang2022efficient, ning2023hoiclip, chao:wacv2018} aims to detect humans and objects in images and classify the interactions between them. While much progress has been made in HOI detection recently, existing approaches primarily focus on a closed-world setting with a fixed number of HOI classes. 

% \newtext{The task of human-object interaction (HOI) detection \cite{gkioxari2018detecting, chao:wacv2018, zhang2022efficient, ning2023hoiclip, zhang2023exploring} aims to detect humans and objects in images and recognize the interactions between them, which is advantageous in real-world autonomous systems such as self-driving vehicles and collaborative robots\cite{10026602,pmlr-v229-mascaro23a}. Recent advancements in HOI detection have been notable, yet most existing approaches are tailored to a closed-world setting with a fixed number of HOI classes. Despite the strong performance of open-vocabulary HOI detectors that leverage linguistic knowledge obtained from vision-language pre-training \cite{radford2021learning,li2023blip}, as demonstrated in \cite{zheng2023open, Yuan2022RLIP}, their capabilities of HOI detection remain confined to the categories explicitly covered by their linguistic vocabularies.}

\begin{figure}[t]
  \centering
   \includegraphics[width=1.0\linewidth]{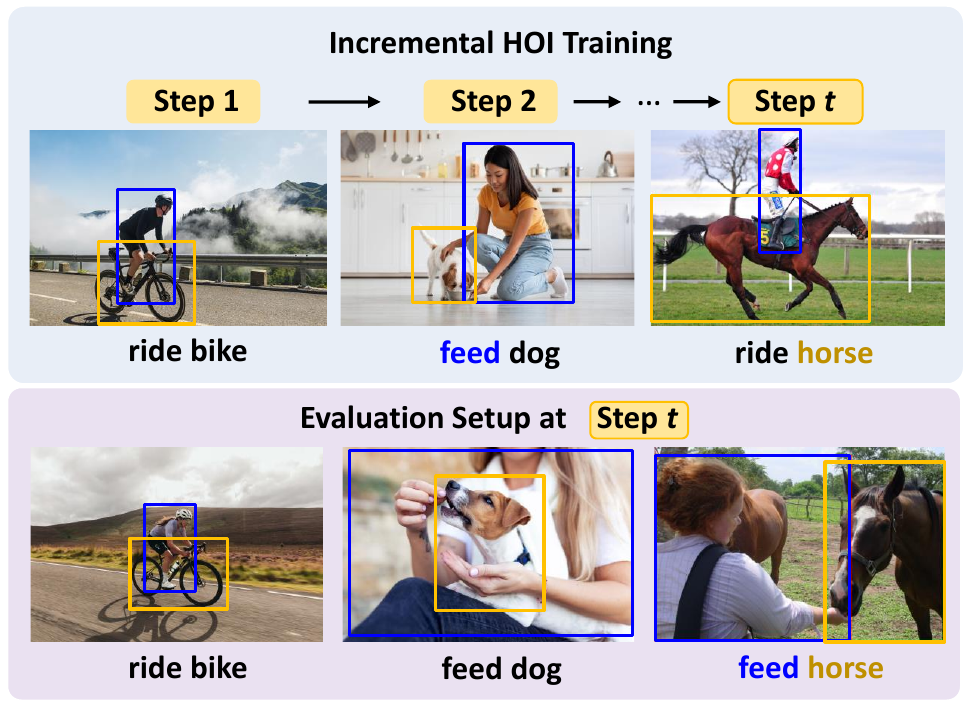}

   % \caption{The training and evaluation of IHOID. The model is trained in multiple phases and learns different object-relation pairs in each phase. During the evaluation, the model should not only detect HOIs learned in the previous and current phases such as \texttt{ride bike} and \texttt{feed dog}, but also overcome interaction drift of \texttt{ride} caused by subsequent \texttt{ride horse}. Additionally, the model is required to recognize the zero-shot HOI \texttt{feed horse}, a combination of previously learned relations and objects.}
   \caption{Training and evaluation of IHOID. The model learns object-relation pairs incrementally and must detect past and new HOIs, mitigate interaction drift, and recognize zero-shot HOIs.}
   \label{fig:adv}
   % \vspace{-10pt}
\end{figure}

Human-object interaction (HOI) detection \citep{gkioxari2018detecting, chao:wacv2018, zhang2022efficient, ning2023hoiclip, zhang2023exploring} involves identifying humans and objects within images and recognizing the interactions between them. This capability holds significant promise for real-world applications such as self-driving vehicles and collaborative robots \citep{10026602,pmlr-v229-mascaro23a}. While recent advancements in HOI detection have been notable, the majority of existing approaches are tailored to closed-world scenarios, where a fixed number of HOI classes are predefined. Despite the impressive performance demonstrated by open-vocabulary HOI detectors \citep{zheng2023open, Yuan2022RLIP}, which utilize linguistic knowledge acquired from vision-language (VL) pre-training \citep{radford2021learning,li2023blip}, their ability to detect HOIs remains limited to the categories explicitly covered by their linguistic vocabularies.

However, in open-world and dynamic environments, it is required to understand long-term human behavior with personalized or task-specific interactions that are hard to pre-define. For instance, home service robots should continually learn to adapt to the evolving actions of users. Besides, in sensitive settings like hospitals, historical data access is restricted due to privacy concerns \citep{babakniya2024data}. 
Consequently, it is highly desirable to endow agents with a human-like capacity for incremental learning \citep{LESORT202052, NEURIPS2022_c5843794}, allowing them to seamlessly integrate new HOI concepts into their knowledge base without the risk of forgetting previously learned ones and without the need to reference past data.
% 具体任务设定 in an incremental learning setting 

In this work, we aim to tackle this problem by introducing an \textit{incremental human-object interaction detection} (\textbf{IHOID}) setup, where the HOI model is trained to progressively detect an increasingly larger set of interactions between humans and a fixed set of familiar objects\footnote{This problem setting reflects a usual daily living or working environment where novel objects often rarely appear but new interactions need to be identified.}. 
Additionally, due to the compositional nature of HOIs, the model should also generalize well to zero-shot object-relation combinations \citep{hou2021fcl, chao:wacv2018, hou2020visual}.
As illustrated in Fig.~\ref{fig:adv}, the model learns the interaction \texttt{feed dog} earlier and incrementally learns new interactions like \texttt{ride horse} at a later time phase, so the model should naturally recognize the novel combination \texttt{feed horse} during evaluation. 

\begin{figure*}[!t]
  \centering
   \includegraphics[width=0.9\linewidth]{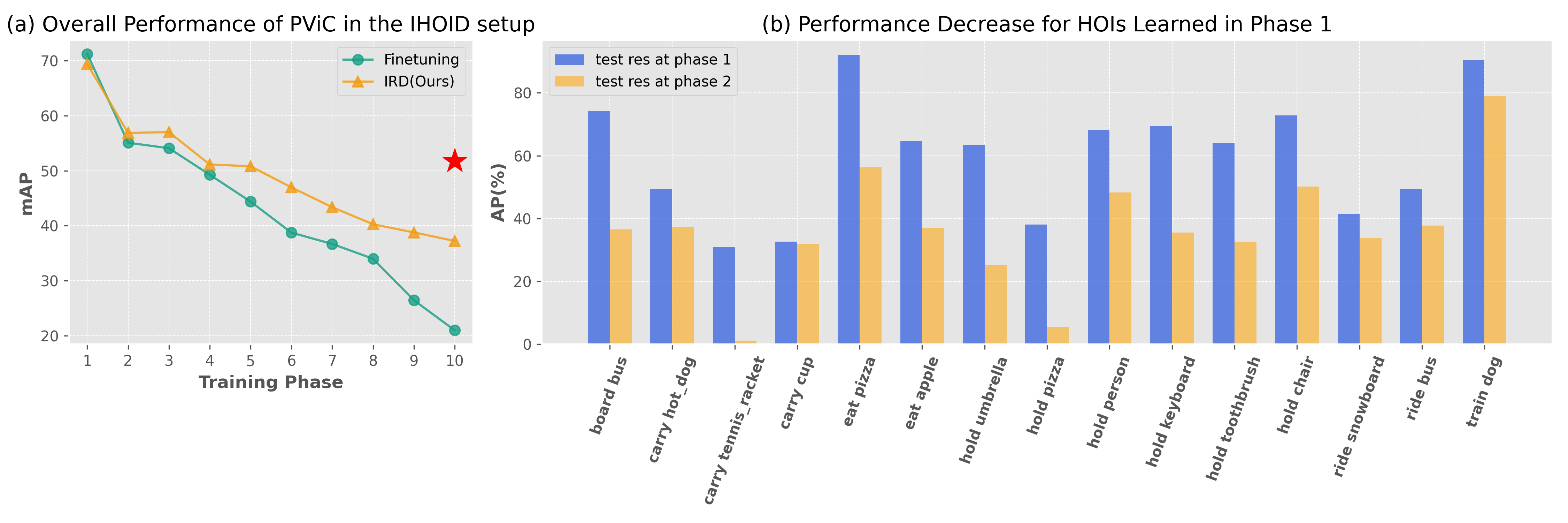}
   % \vspace{-10pt}
    \caption{\textbf{(a)} Performance degradation of the SOTA HOI detector PViC in the IHOID setup: The yellow plot shows the incremental training performance of PViC on our partitioned HICO-DET dataset. The red star denotes the performance achieved by PViC under a joint training setup with an identical dataset, which serves as the upper bound for the model trained in the IHOID setup. \textbf{(b)} Demonstration of interaction drift: The statistics show the APs of HOI categories which are related to the same relation categories that occur across training phase 1 and phase 2. The APs of these categories suffer from obvious decreases.}

   \label{fig:sta}
   \vspace{-10pt}
\end{figure*}

% \textcolor{red}{However, IHOID introduces unique challenges beyond standard class incremental learning. In addition to catastrophic forgetting, where newly learned interactions overwrite previous knowledge, two key issues arise: interaction drift, where learning new HOIs disrupts existing relational representations, and zero-shot HOI generalization, where the model must infer novel interactions across disjoint learning phases. We elaborate the challenge analysis in sec3, as shown in Fig.~\ref{fig:sta}, even state-of-the-art HOI detectors like PViC suffer performance degradation under incremental training.}

However, IHOID introduces unique challenges beyond standard class incremental learning. In addition to catastrophic forgetting, two key issues arise. First, \textit{interaction drift} occurs when learning new HOIs alters the representations of previously acquired interactions that share the same relation category (e.g., \texttt{ride} in step $1$ and step $t$ in Fig.~\ref{fig:adv}). This is due to the model’s excessive reliance on object-specific features rather than learning robust relational representations. Second, \textit{zero-shot HOI generalization} requires the model to infer novel interactions across disjoint learning phases, where objects and relations appear at different times with limited contextual exposure. 
To address the challenges of incremental HOI detection, we propose an exemplar-free Incremental Relation Distillation (\textbf{IRD}) framework, which mitigates catastrophic forgetting, counteracts interaction drift, and enhances zero-shot generalization.
IRD separates the learning of objects and relations, reducing the dependency of relation representations on specific objects. Also, to achieve robust and adaptable relation learning, we introduce two novel distillation strategies: (1) Concept Feature Distillation (CFD) enforces relation consistency across object contexts, ensuring that interactions like \texttt{ride} remain invariant whether paired with \texttt{bicycle} or \texttt{horse}.
(2) Momentum Feature Distillation (MFD) smooths knowledge transitions across learning phases, preserving discriminative relation features while integrating new HOIs.

We validate our approach by extensive comparison with prior incremental learning and zero-shot HOI detection methods on two widely used HOI datasets: HICO-DET \citep{chao:wacv2018} and V-COCO \citep{gupta2015visual}. The experimental results and ablation study show that our method outperforms other approaches in tackling catastrophic forgetting and interaction drift and has better generalization on zero-shot HOIs.

Our main contributions can be summarized as follows:
\begin{itemize}
  \item We propose the incremental learning setting for human-object interaction detection (IHOID), which focuses not only on the catastrophic forgetting of HOI classes but also on the model's robustness to interaction drift and generalization ability on zero-shot HOI combinations.
  % \item We propose a novel knowledge distillation method CCD to mitigate the forgetting problem caused by new classes, while enabling the model to learn consistent representations for the same actions during the continual learning process to deal with interaction shift and unseen HOI combinations.
  \item To tackle the challenges introduced by IHOID, we propose an exemplar-free incremental relation distillation framework that independently supervises the learning of objects and relations and focuses on learning robust and invariant relation representations via two complementary distillation strategies, namely CFD and MFD. 
  \item We conduct extensive experiments on partitioned HICO-DET and V-COCO, demonstrating that our method outperforms the SOTA baselines under the aforementioned two new challenges along with catastrophic forgetting.
\end{itemize}

\section{Related Works}

\subsection{Incremental Learning}
In class incremental learning (CIL) \citep{li2017learning, rebuffi2017icarl, wu2019large, goswami2024fecam, asadi2023prototype}, models sequentially learn new classes from incoming data batches, a crucial capability for agents adapting to evolving environments \citep{ayub2024interactive}. However, this process often leads to catastrophic forgetting \citep{kirkpatrick2017overcoming}, where previously learned knowledge is overwritten by new information. Existing CIL approaches fall into three categories: (1) Dynamic architecture methods \citep{xie2022general, yan2021dynamically, douillard2022dytox, hu2023dense} expand model structures to accommodate new classes. (2) Memory-based methods \citep{sarfraz2023error, bang2021rainbow, rebuffi2017icarl, wang2021ordisco, shi2024unified, bellitto2025saliency} store exemplars and use memory replay for continual learning. (3) Regularization-based methods \citep{li2017learning, douillard2020podnet, shi2021continual, song2023ecotta, asadi2023prototype} constrain weight updates to mitigate forgetting.  
In addition to these, researchers have delved into incremental learning for perception tasks like object detection \citep{liu2023augmented,feng2022overcoming} and segmentation \citep{chen2024saving,xiao2023endpoints,oh2022alife,cermelli2022incremental}, where \citet{liu2023continual, liu2023augmented} proposes task-specific designs that leverage memory and distillation losses to optimize learning.
Unlike standard CIL, where forgetting mainly occurs when introducing new categories, IHOID presents the additional challenge of interaction drift, which cannot be effectively addressed by existing CIL methods designed for object-centric tasks.

\subsection{Standard and Zero-Shot HOI Detection}  
Human-object interaction (HOI) detection \citep{chao:wacv2018, gupta2015visual, gkioxari2018detecting, li2020hoi, li2019transferable} is crucial for understanding structured scenes by capturing both objects and their interactions. Traditional methods operate in a closed-world setting, relying on predefined categories and static datasets. These approaches can be categorized into two-stage models \citep{gao2018ican, li2019transferable, wang2020contextual, zhang2023exploring}, which first detect objects before inferring interactions, and one-stage models \citep{chen2021reformulating, tamura2021qpic, liao2022gen}, which predict HOI triplets directly.  
To extend beyond fixed categories, recent open-vocabulary HOI detection methods \citep{yuan2023rlipv2, zhao2023unified} integrate vision-language (VL) models \citep{li2023blip} or large language models \citep{openai2023gpt}. However, these approaches remain constrained by the vocabulary within pre-trained datasets. Zero-shot HOI detection further generalizes to unseen HOIs through compositional learning \citep{hou2020visual, hou2021fcl} or VL pre-training \citep{Yuan2022RLIP, zheng2023open}.  

Furthermore, our IHOID setup challenges models to continuously expand their HOI knowledge in an open-ended manner. It not only requires models to learn from a continuously arriving data stream but also to naturally generalize to zero-shot HOI combinations. This setup better aligns with real-world learning, where interactions emerge dynamically rather than being predefined.

% Unlike zero-shot learning, which infers unseen interactions from existing knowledge, incremental HOI detection (IHOID) requires models to continually acquire and retain new HOIs. Additionally, IHOID introduces interaction drift, where learning new interactions disrupts previously learned ones. Our work addresses these challenges by developing an incremental learning framework that ensures continual adaptation while maintaining relational consistency.

\section{Problem and Challenge}
\subsection{Problem Formulation}
\label{sec:formulation}
% In the IHOID setup, we aim at incrementally learning HOI categories, while maintaining the generalization ability to unseen HOI combinations.
In the IHOID setup, our objective is to address the challenges of mitigating catastrophic forgetting of HOI classes while simultaneously preserving the model's robustness against interaction drift and enhancing its generalization capabilities for unseen HOI combinations. 
In the problem formulation, the HOI detector is subjected to incremental learning over a total of $T$ training phases. 
During each phase $t\in\{1,\cdots,T\}$, the model is exposed to only a subset of annotations corresponding to specific HOI categories.

We formally define the training set as $\mathcal{D}=\{(I,y)\}$, where $I$ denotes the images and $y$ represents the corresponding HOI annotations. Within the annotations $y$, we introduce $\mathcal{C}=\{C_i\}_{i=1}^{N_c}$ as the set of human-object interactions, $\mathcal{O}=\{O_j\}_{j=1}^{N_o}$ as the set of objects, and $\mathcal{R}=\{R_k\}_{k=1}^{N_r}$ as the sets of relations. Here, $N_c, N_o,$ and $N_r$ denote the counts of HOI, object, and relation categories, respectively. Each HOI category $C_i$ is composed of an object category $O_j$ paired with a relation category $R_k$.
% The set of fixed objects is denoted as $\mathcal{O}=\{O_j\}_{j=1}^{N_o}$, while the incrementally arriving HOI and relation categories are represented by $\mathcal{C}=\{C_i\}_{i=1}^{N_c}$ and $\mathcal{R}=\{R_k\}_{k=1}^{N_r}$, respectively.
% Given a fixed familiar object set $\mathcal{O}=\{O_j\}$, we let $\mathcal{C}=\{C_i\}$ and $\mathcal{R}=\{R_k\}$ be the set of incrementally arriving interaction categories and action categories, respectively. 

To  establish the framework for the IHOID task, we partition the dataset and HOI categories into $T$ disjoint subsets, denoted as $\mathcal{D}=\mathcal{D}_1 \cup \cdots \cup \mathcal{D}_T $ and $\mathcal{C}=\mathcal{C}_1 \cup \cdots \cup \mathcal{C}_T$, respectively, assigning one to each training phase. In each phase $t$, we filter samples $\{(I,y)\} \subseteq \mathcal{D}_t$ such that $y$ comprises only the HOI annotations belonging to $\mathcal{C}_t$. Upon completion of phase $t$, the training switches to phase $t+1$, introduing the model to a different set of images $\mathcal{D}_{t+1}$ and corresponding HOI annotations $\mathcal{C}_{t+1}$. The specific distribution of HOI categories across phases is elaborated in Section~\ref{sec:exp_set}.

% {Notably, the IHOID task naturally preserves the multi-label nature of HOI, because at each learning phase, the model must predict multiple relation categories associated with the interaction for every detected human-object box pair, as long as these categories are present in the current or previous training phases. For instance, when humans ride bicycles, they also \textit{sit on} the bicycle, \textit{straddle} the bicycle, and \textit{hold} the bicycle, so the model needs to predict all these categories simultaneously.}

Notably, the IHOID task inherently retains the multi-label nature of HOI detection. At each learning phase, the model must predict multiple relation categories associated with each detected human-object box pair, provided these categories have been encountered in the current or previous training phases. For instance, when a person rides a bicycle, he may also \textit{sit on}, \textit{straddle}, and \textit{hold} the bicycle, requiring the model to predict all these interactions simultaneously.

\begin{figure*}[!t]
    \centering
    \includegraphics[width=0.85\linewidth]{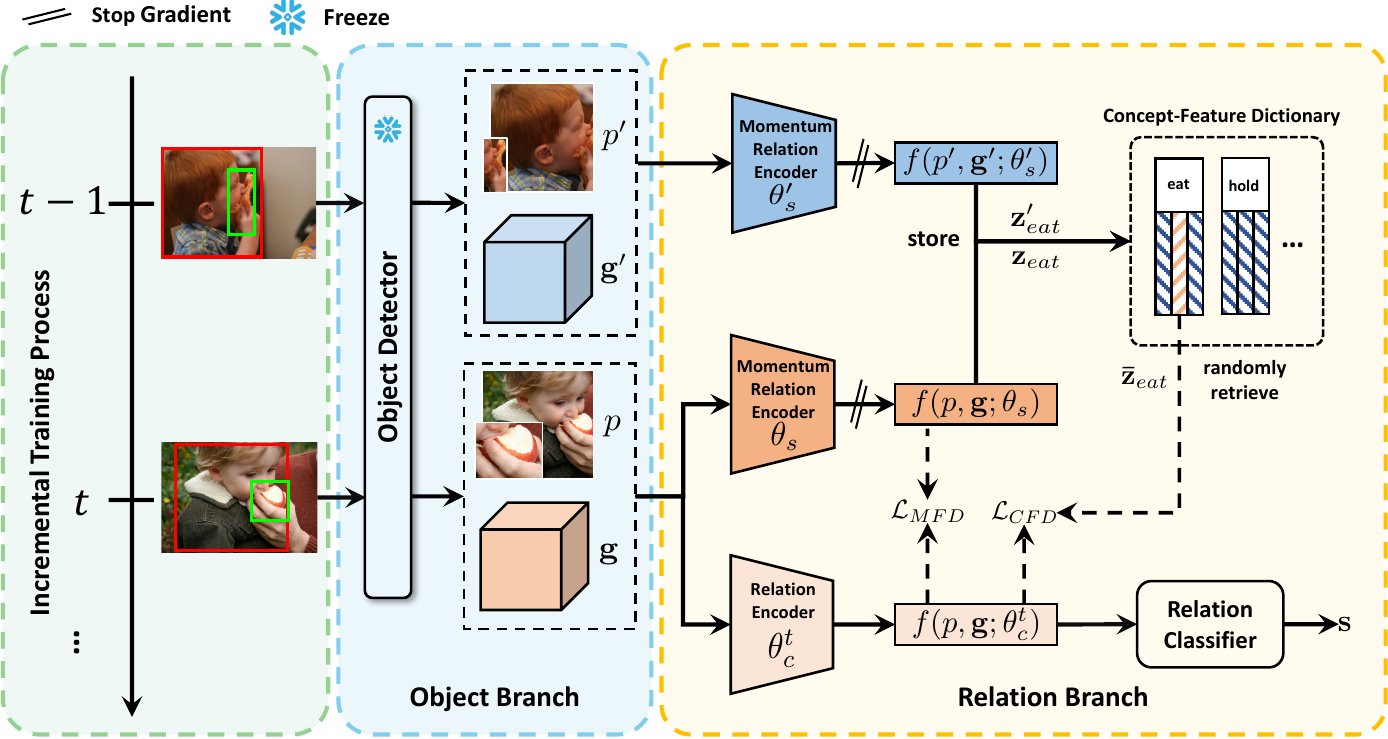}
    \caption{
    % The pipeline of our proposed SIR framework. A pretrained object detector produces human and object boxes with their detection scores, and boxes are 
    The pipeline of our relation representation learning framework. At each training phase $t$, the object branch outputs the box pair information $p$ and the global image feature $\mathbf{g}$. These are then fed into the relation branch, where a momentum teacher processes them to produce the reference relation feature ${\mathbf{z}}=f(p,\mathbf{g};\theta_s)$, subsequently stored in the concept-feature dictionary. Concurrently, the current encoder takes the same input and yields $f(p,\mathbf{g};\theta_c^t)$, facilitating the computation of distillation losses $\mathcal{L}_{MFD}$ and $\mathcal{L}_{CFD}$ with ${\mathbf{z}}$ and the invariant relation feature $\mathbf{\bar{z}}$ randomly retrieved from the dictionary, respectively. 
    % Additionally, the relation logits $\Tilde{\mathbf{s}}$ and $\mathbf{s}$, produced by the last step and current step relation classifiers, are used to compute $\mathcal{L}_{CDD}$.
    }
    \label{fig:model}
    \vspace{-15pt}
\end{figure*}
\subsection{Challenge Analysis}
\label{sec:challenge}
The IHOID setup presents challenging problems for exploration, as illustrated in Fig.~\ref{fig:sta}\textcolor{iccvblue}{a}, where even the state-of-the-art HOI detector PViC~\citep{zhang2023exploring} experiences a degradation in performance during incremental training.
This setting not only faces the widely acknowledged difficulty of catastrophic forgetting in CIL, but also introduces two novel challenges.

First, the compositional nature of HOI classes leads to a unique challenge we term \textit{interaction drift}. Since multiple HOI classes share the same relation category, learning a new interaction may interfere with previously learned ones. For instance, after learning \texttt{ride bike}, the subsequent acquisition of \texttt{ride horse} may overwrite or distort the learned representation of \texttt{ride bike}, even though both interactions fall under the same relational concept. This issue primarily arises due to the model’s excessive dependence on object-specific features rather than learning robust relation representations. The impact of this phenomenon is quantified in Fig.~\ref{fig:sta}\textcolor{iccvblue}{b}.  

Second, IHOID differs fundamentally from zero-shot HOI learning, where unseen interactions are inferred from pre-existing knowledge in a joint training framework \citep{hou2021fcl, hou2021atl}. In our setting, objects and relations associated with zero-shot cases emerge at different time phases, and the model is exposed to only a partial dataset at each phase. This fragmented exposure limits the model’s ability to generalize to novel HOI compositions. As shown in Fig.~\ref{fig:adv}, the test example \texttt{feed horse} demonstrates the difficulty of generalization when learning occurs incrementally rather than holistically.

% \vspace{-5pt}
\section{Methods}
% \vspace{-5pt}
We introduce a novel Incremental Relation Distillation (IRD) framework to overcome the challenges for incrementally learning the compositional object-relation classes.
In the following subsections, we first present the overview of model architecture in Sec.~\ref{sec:model-arch}. In
Sec.~\ref{sec:sir}, we elaborate on the proposed method which facilitates the learning of relation representations through distillations. Finally, we conclude the training objective functions of this framework in Sec.~\ref{sec:model-learning}.

\subsection{Model Architecture}
\label{sec:model-arch}

We propose a model architecture that disentangles the learning of object and relation categories, allowing the model to learn relation representations independent of object-specific features.

As shown in Fig.~\ref{fig:model}, the model consists of two primary components: an object branch and a relation branch. In the object branch, for an input image $I$, we utilize a pre-trained object detector based on the H-Deformable DETR \citep{jia2023detrs} architecture to generate a global image feature $\mathbf{g}$ and a set of object detection results. Non-maximum suppression and thresholding are subsequently applied, leaving a smaller result set $\{d_i\}^n_{i=1}$, where $d_i=(\mathbf{b}_i, s_i, c_i, \mathbf{x}_i)$ consists of the box coordinates $\mathbf{b}_i \in \mathbb{R}^4$, the confidence score $s_i \in [0,1]$, the predicted object class $c_i \in \mathcal{O}$, and the object feature $\mathbf{x}_i$.
The output boxes are paired as human-object candidates, forming the set  $\mathcal{P}=\{p=(\mathbf{x}_i,\mathbf{x}_j, \mathbf{b}_i,\mathbf{b}_j)~\vert~i \neq j, c_i=\text{human}\}$. In the relation branch, together with the global feature $\mathbf{g}$, $p$ is taken as input to the relation encoder $f$ parameterized by $\theta$, producing the relation representation $f(p,\mathbf{g};\theta)$ for the box pair $p$, and finally being fed to the relation classifier to predict the relation logits $\mathbf{s}$. To fully leverage the information from the pre-trained object detector, we integrate the object confidence scores into the final score computation of each human-object pair. The final score of $p$ is formulated as:
\begin{equation}
    \tilde{\mathbf{s}} = ({s_i}\cdot{s_j})^{1-\lambda} \cdot \sigma(\mathbf{s})^\lambda
    \label{eq:eq1}
\end{equation}
where $\lambda$ is a constant to suppress overconfident objects \citep{zhang2022efficient} and $\sigma$ is the sigmoid function.
% The predictions for objects are generated from the object detector. 
The training loss $\mathcal{L}_{rel}$ for this architecture is the focal loss \citep{lin2017focal} on the relation classification, which deals with the imbalance between positive and negative examples.

% For the relation encoder, we follow the interaction head of the state-of-the-art HOI detector PViC \cite{zhang2023exploring} to design its structure, and we also follow \cite{hou2019learning} to introduce cosine normalization to the vanilla softmax function in the relation classifier so that the model can be less biased to new classes.
For the design of the relation encoder, we adopt the architecture of the interaction head from the state-of-the-art HOI detector PViC \citep{zhang2023exploring}. Additionally, To mitigate the model's bias towards new classes during the incremental learning process, we incorporate cosine normalization into the standard softmax function within the relation classifier, as introduced in \citet{hou2019learning}. 
%in incremental learning, which is naturally a disentangled model architecture. However, it is worth mentioning that our framework can also be applied to other HOI detectors with a similar disentanglement design \cite{shen2018scaling, hou2020visual}.
Furthermore, given that the object categories $\mathcal{O}$ are known beforehand, we propose freezing the object detector, which has been pre-trained on all object categories within the dataset. This approach allows us to concentrate on advancing the learning capabilities of the relation branch.

\subsection{Invariant Relation Distillation}
\label{sec:sir}

In this section, we present two complementary distillation strategies: Momentum Feature Distillation (MFD) and Concept Feature Distillation (CFD), implemented via a momentum teacher and a novel concept-feature dictionary, respectively. These strategies ensure stable and transferable relation representations, preserving semantic integrity across incremental learning phases while adapting to new interactions. The following subsections detail their implementation and integration into our framework, along with the corresponding loss functions.
\subsubsection{Momentum Feature Distillation}
% While previous knowledge distillation (KD) methods have utilized models from old steps as teachers to alleviate forgetting, we propose using a more stable model as the teacher in our approach. This stable model provides relatively consistent action representations as a reference for the current model's learning process. This is achieved by maintaining old step knowledge while performing minimal adaptation and fine-tuning using the current step data. 
% This ensures that the model's output for old classes exhibits semantically consistent representations with the old step model.
The abrupt shift in data distributions between phases causes conventional knowledge distillation \citep{douillard2020podnet,hou2019learning,li2017learning} to struggle, as it merely transfers knowledge from a static model from the last phase that fails to adapt to the nuances of new data. To address this fundamental limitation, we introduce Momentum Feature Distillation (MFD), a dynamic knowledge transfer mechanism that was first used in unsupervised learning \citep{he2020momentum,caron2021emerging}, to create a balanced bridge between preserving past knowledge and accommodating new concepts.
% \sout{While previous knowledge distillation methods \citep{douillard2020podnet,hou2019learning,li2017learning} in continual learning only used the model from the last phase as the teacher, in our approach, we employ a more robust momentum teacher which was firstly used in unsupervised learning \citep{he2020momentum,caron2021emerging}. The teacher model retains the knowledge from previous phases and mildly adapts to the data in the current phase. }

Specifically, in addition to maintaining a frequently changing current model $\theta_c^t$ at phase $t$, we keep a model $\theta_s$ as the momentum teacher, which remains detached from the training process. At each iteration, the current model $\theta_c^t$ adapts to the target distribution and simultaneously updates the  model $\theta_s$ using exponential moving weighted average:
\begin{equation}
    \theta_s = m \theta_s + \texttt{sg}[(1-m) \theta_c^t]
    \label{eq:eq2}
\end{equation}
where \texttt{sg} is the stop-gradient operation and $m$ is the momentum value.
% , and the stable model $\theta_s$ is updated to be a weighted average of its previous value and the current model's value.
% \sout{By introducing this momentum teacher, we ensure a stable reference for the model's learning process, preserving old knowledge while allowing soft adaptation to new data distributions.}
For each human-object pair $p$, the MFD loss is formulated as:
\begin{equation}
\mathcal{L}_{MFD} = \left\| f(p,\mathbf{g};\theta_s) - f(p,\mathbf{g};\theta_c^t) \right\|_2^2
\label{eq:eq4}
\end{equation}
where $f(p,\mathbf{g};\theta_s)$ and $f(p,\mathbf{g};\theta_c^t)$ are the relation representations obtained from the momentum teacher and the current model, respectively. This dynamic balancing act enables our model to incrementally adapt to new interaction classes while preserving a stable representational space for previously learned concepts.

\subsubsection{Concept Feature Distillation} 

Sec.~\ref{sec:challenge} analyzes the problematic dependencies between relations and specific objects. Based on this, we propose a concept-feature dictionary that systematically captures invariant relation features across diverse object contexts. This dynamic dictionary ensures that relations maintain consistent semantic properties regardless of their object pairings—e.g., the action \texttt{ride} exhibits fundamental patterns whether applied to a \texttt{bicycle} or a \texttt{horse}.  
Structured as separate queues for each relation concept, the dictionary enables efficient storage and retrieval of relation prototypes, allowing the model to preserve relation consistency, mitigate interaction drift, and generalize to unseen combinations in an incremental learning scenario. Building upon this dictionary, we introduce the Concept-Feature Distillation (CFD) loss, which fully exploits its structure to enhance the learning of invariant relation representations. The following subsections detail the design of both the dictionary and the loss function.
\vspace{-5pt}
\paragraph{Concept definition:} In our context, a concept represents a relation category, although it can be adapted to other entities like objects, attributes, or HOI categories in different incremental learning frameworks \citep{feng2022overcoming,michieli2021continual}.
\vspace{-5pt}
\paragraph{Dictionary structure:} For each concept, the dictionary maintains a queue of invariant reference representations. At training phase $t$, let the total number of learned concepts be $N_t$ and the accumulated learned set of concepts up to phase $t$ be $\mathcal{R}_{1:t}=\{R_1,\cdots,R_{N_t}\}$. The dictionary is represented as $\{(R_1,Q_1),\cdots,(R_{N_t},Q_{N_t})\}$, pairing each relation concept $R_i$ with a queue $Q_i$ of capacity $L$.
\vspace{-5pt}
\paragraph{Concepts for box pairs:} When processing one image, we select a subset of candidate box pairs $\mathcal{P}_s$ from the predicted pairs $\mathcal{P}$, ensuring each $p\in\mathcal{P}_s$ has a minimum box-pair Intersection over Union (IoU) of 0.5 with its ground truth. Note that a box pair $p$ may correspond to multiple relation concepts, we define $\mathcal{R}_{p} \subseteq \mathcal{R}_{1:t}$ as the set of concepts related to $p$. 
% For each relation concept $r\in\mathcal{R}_p$, the associated relation representation is enqueued in its corresponding queue within the dictionary.
\vspace{-5pt}
\paragraph{Storage and retrieval:} For any pair $(p, \mathcal{R}_{p})$, we select a concept $R\in\mathcal{R}_{p}$ and randomly retrieve a relation feature $\mathbf{\bar{z}}$ from its corresponding queue $Q$, which is then utilized to compute the CFD loss. Concurrently, the box pair $p$ is processed through the teacher network $\theta_s$ to generate a new relation feature ${\mathbf{z}}= f(p,\mathbf{g}; \theta_s)$, which is subsequently enqueued into $Q$. If $Q$ reaches capacity, the oldest feature is removed.
\vspace{-5pt}
\paragraph{Initialization and update:} Initially, the dictionary is empty. For a new concept $R\in\mathcal{R}_{p}$ which is absent in the dictionary, a new entry $(R, Q)$ is created, and the feature ${\mathbf{z}}$ is added to $Q$ without retrieval. The dictionary undergoes continual updates at each training iteration, enabling the persistent growth and refinement of reference features.
\vspace{-5pt}
\paragraph{Distillation Strategy:} Building upon our concept-feature dictionary, we introduce CFD loss, a novel distillation strategy that explicitly encourages the learning of object-invariant relation representations. For each box pair $p$, the CFD loss is defined as
\begin{equation}
    \mathcal{L}_{CFD}=\Vert f(p,\mathbf{g};\theta_c^t)-\mathbf{\bar{z}}\Vert_2^2
    \label{eq:eq3}
\end{equation}
where $\mathbf{\bar{z}}$ is the invariant relation representation retrieved from the concept-feature dictionary.

% \subsubsection{Concept Feature Distillation}
% To encourage the model to learn invariant characteristics of the same relation category across different samples, we introduce the CFD loss.
% For each box pair $p$, the CFD loss is defined as
% \begin{equation}
%     \mathcal{L}_{CFD}=\Vert f(p,\mathbf{g};\theta_c^t)-\mathbf{\bar{z}}\Vert_2^2
%     \label{eq:eq3}
% \end{equation}
% where $\mathbf{\bar{z}}$ is the invariant relation representation retrieved from the concept-feature dictionary. 
% It corresponds to the same relation concept as the one involved in box pair $p_k$.
% The goal is to minimize the Euclidean distance between the relation representation obtained by the current model and the invariant representation $\mathbf{z}$ and encourage the model to learn the invariant characteristics across different samples.
% This promotes semantic consistency among different samples that share the same action concept.

% \subsubsection{Momentum Feature Distillation}

% To simultaneously learn HOI classes with new relation categories while retaining the knowledge from previous phases, we develop Momentum Feature Distillation (MFD) to learn stable relation representations. For each box pair $p$, the MFD loss is computed as
% \begin{equation}
% \mathcal{L}_{MFD} = \left\| f(p,\mathbf{g};\theta_s) - f(p,\mathbf{g};\theta_c^t) \right\|_2^2
% \label{eq:eq4}
% \end{equation}
% where $f(p,\mathbf{g};\theta_s)$ and $f(p,\mathbf{g};\theta_c^t)$ are the relation representations obtained from the momentum teacher and the current model, respectively.

\subsubsection{Concept Distribution Distillation}
In addition to the proposed two distillations, we employ a classic technique known as Concept Distribution Distillation (CDD) \citep{li2017learning} to prevent the forgetting of the classifier. For each box pair $p$, with a maintained model $\theta_c^{t-1}$ from the last phase, this distillation loss is defined as follows:
\begin{equation}
    \mathcal{L}_{CDD}=-\sum\limits_{i=1}^{N_{t-1}}\mathbf{q}_i^{t-1}\log\mathbf{q}_i^t
    \label{eq:eq5}
\end{equation}
where $\mathbf{q}_i^t=\frac{e^{\mathbf{s}_i^t/T}}{\sum_{j=1}^{N_r^{t-1}}{e^{\mathbf{s}_j^t/T}}}$, $\mathbf{q}_i^{t-1}=\frac{e^{\mathbf{s}_i^{t-1}/T}}{\sum_{j=1}^{N_r^{t-1}}{e^{\mathbf{s}_j^{t-1}/T}}}$, $N_r^{t-1}$ is the number of learned relation categories until the end of phase $t-1$, $\mathbf{s}_i^t$ is the $i^{th}$ element in the logits $\mathbf{s}^t$ given by the current phase model $\theta_c^t$, $\mathbf{s}_i^{t-1}$ is the $i^{th}$ element in the logits $\mathbf{s}^{t-1}$ given by the last phase model $\theta_c^{t-1}$, and $T$ is the temperature set as $T=1$ by default.

% Additionally, we allow the model to use a memory \cite{riemer2018learning} with limited size to store and replay exemplar samples from previous steps.

\subsection{Training Objectives}
\label{sec:model-learning}
In the training stage, the total loss $\mathcal{L}_{total}$ is the weighted sum of four components calculated over all box-pair candidates: the standard relation classification loss $\mathcal{L}_{rel}$ illustrated in Sec.~\ref{sec:model-arch}, CDD loss, CFD loss, and MFD loss. $\mathcal{L}_{total}$ is thereby formulated as
% \vspace{-5pt}
\begin{equation}
    \mathcal{L}_{total} = \sum_{p\in\mathcal{P}_s}\left(\mathcal{L}_{rel} + \alpha_0\mathcal{L}_{CDD} + \alpha_1\mathcal{L}_{MFD} + \alpha_2\mathcal{L}_{CFD}\right)
    \label{eq:eq6}
    % \vspace{-5pt}
\end{equation}
where $\alpha_0, \alpha_1, \alpha_2$ are tunable hyperparameters used to balance the contribution of each loss term.

\section{Experiments}

We conduct a series of experiments to verify the effectiveness of our method. In this section, we first introduce the experiment setup in Sec.~\ref{sec:exp_set}. Then we show our experimental results in Sec.~\ref{sec:exp_res}, followed by the ablation study in Sec.~\ref{sec:exp_abl}.
% In the end, we show the t-SNE visualization of the relation representations to provide more insights into our method.

% In this section, we examine several baseline models in the context of incremental learning and zero-shot HOI on the proposed task. Our evaluation primarily centers around the models' robustness against interaction drift and their capability to generalize to unseen HOI combinations.
% Subsequently, we conduct a comparative analysis between our model and the aforementioned strong baselines, demonstrating its effectiveness within these specific settings.

% \vspace{-10pt}
\subsection{Experiment Setup}
\label{sec:exp_set}

% \vspace{-5pt}
\subsubsection{Baselines}
% \textcolor{red}{The baselines we compare with include incremental learning stategies and methods for zero-shot HOI detection, which can be transferred to our setup, allowing us to showcase the advantages of our approach in this task comprehensively.
% Firstly, we investigate whether the challenges of this new problem can be effectively addressed by several classical and SOTA class incremental learning methods which can be transferred to our setup, namely LwF \cite{li2017learning}, PODNet-flat \cite{douillard2020podnet}, PCR \cite{lin2023pcr}, and PRD \cite{asadi2023prototype} in comparison to our approach.
% Additionally, we examine the adaptability of General-Inc \cite{xie2022general}, a proposed method for solving the general incremental learning problem, to our task. 
% Lastly, for a comprehensive and fair comparison, we apply zero-shot detection methods VCL \cite{hou2020visual} and SCL \cite{hou2022discovering} to our model architecture, along with General-Inc which performs best among other previous methods in our setup, as baselines for zero-shot HOI detection in the IHOID setup.
% To match the exemplar-free setup of IHOID, all adopted CIL baselines except PCR are non-exemplar methods. For PCR, we remove its memory component (the rest part is referred as PCR-MEM0) during the experiment for fair non-examplar comparisons.
% }

The baselines we compare with encompass incremental learning strategies and zero-shot HOI detection methods. 
% This allows for a detailed evaluation of our approach's effectiveness in this task.
We first evaluate the capability of several classical and SOTA class incremental learning methods to tackle the unique challenges presented by our problem. The methods considered for comparison include LwF \citep{li2017learning}, PODNet-flat \citep{douillard2020podnet}, PCR \citep{lin2023pcr}, and PRD \citep{asadi2023prototype}, all adapted to fit our experimental setup.
Additionally, we explore the applicability of General-Inc \citep{xie2022general}, a proposed method for general incremental learning challenges, in the context of IHOID. For a comprehensive evaluation, we also apply zero-shot detection methods VCL \citep{hou2020visual} and SCL \citep{hou2022discovering} to our model architecture alongside General-Inc, which exhibited good performance compared with prior methods in our setup, as baselines for zero-shot HOI detection. Moreover, we train our HOI detector on the entire training set (joint training) and acquire the upper-bound performance for reference. Besides, To ensure consistency with the exemplar-free nature of the IHOID task, all class incremental learning (CIL) baselines, except PCR, are non-exemplar methods. For PCR, we omit its memory component in our experiments to maintain fair comparisons among non-exemplar approaches.
Details on adapting these baselines to the IHOID setup are provided in Suppl.

% The modeling of domains in this problem is similar to the learning of interactions with the same relation and different objects.

% \vspace{-15pt}
\subsubsection{Datasets}
To investigate the IHOID setting, we conduct experiments on two widely used HOI datasets HICO-DET \citep{chao:wacv2018} and V-COCO \citep{gupta2015visual}.
We perform preprocessing on them, including removing the \texttt{no interaction} category in HICO-DET and excluding four body motion categories and the \texttt{point instr} category in V-COCO following \citet{zhang2022efficient}. Specifically, any HOI and its corresponding bounding box annotations related to these relation categories are removed, and images lacking annotations after the removal are also discarded. The detailed statistics of two datasets before and after preprocessing are shown in Suppl.
% We perform preprocessing on the HOI datasets and divide them into multiple splits to support training and testing in different time steps.
% HICO-DET is a large-scale HOI detection dataset consisting of 33,601 training images, 8,528 test images, 80 object types, 116 actions, and 520 interaction types. The dataset includes 102,450 annotated human-object pairs in the original training set and 29,110 in the test set.
% V-COCO, on the other hand, is relatively smaller in scale, with 5,400 images in the train-val set and 4,946 images in the test set, with 80 objects, 24 actions, and 287 interaction types.
% For the relation categories in HICO-DET, we remove the \texttt{no interaction} category. In V-COCO, following the previous work on HOI detection \cite{zhang2022efficient}, we exclude four body motion categories and the \texttt{point instr} category. Specifically, any HOI and its corresponding bounding box annotations related to these relation categories are eliminated. Images lacking annotations after the removal are also discarded.
% We also conduct experiments on a large-vocabulary HOI dataset SWiG-HOI \cite{wang2021discovering}, and details will be presented in Suppl.

% \vspace{-15pt}
\subsubsection{Training Set Partition} When partitioning the training set for each learning phase, we follow the problem formulation guidelines in Sec.~\ref{sec:formulation}.
% and ensured that all objects and relations appeared in the training set. Consequently, 
Object-relation pairs that do not appear during training are considered as unseen HOI combinations, constituting our zero-shot test samples.
% Detailedly, about the split of HOIs for each step, HOI category subsets of different steps have no overlap, namely $\mathcal{C}_i \cap \mathcal{C}_j = \varnothing, i,j \in \{1,\cdots, M\} $.
Specifically, each new HOI class that emerges in training phase $t$ is characterized by the introduction of either a new object or a new relation category not present in previous phases. 
Formally, for $C_i=(O_j, R_k)$ in $\mathcal{C}_t$, either $O_j \notin \mathcal{O}_{1:t-1} $ or $ R_k \notin \mathcal{R}_{1:t-1}$ holds true. 
We partition HICO-DET into 5-phase and 10-phase training subsets, and V-COCO is split into 5-phase subsets.
% whereas V-COCO is only split into 5-phase subsets, as a 10-phase division results in too small subsets for effective training. 
Detailed information on the statistics of the partitions is shown in the Suppl.
% This restriction challenges the model for zero-shot testing.

% \vspace{-15pt}

\begin{table*}[!t]
    % \footnotesize
	\centering
	\caption{Experiment results of our model compared with other incremental learning methods on HICO-DET and V-COCO datasets, specifically preprocessed for the IHOID setup.}
    \resizebox{\textwidth}{!}{
        \renewcommand{\arraystretch}{1.1}
        % \fontsize{20pt}{20pt}\selectfont
    	\begin{tabular}{c|cccccc|cccccc|cccc}
    		\toprule[1pt]
            & \multicolumn{12}{c|}{\textbf{HICO-DET}} & \multicolumn{4}{c}{\textbf{V-COCO}} \\
            & \multicolumn{6}{c|}{$T=5$} & \multicolumn{6}{c|}{$T=10$} & \multicolumn{4}{c}{$T=5$}  \\
    		\multirow{-3}{*}{\textbf{Methods}} & Old & Full & Rare & Non-rare & RID & UC & Old & Full & Rare & Non-rare & RID & UC & Old & Full & RID & UC  \\ \hline
    		Joint (Upper Bound) & - & 51.02 & 42.04 & 53.56 & - & 21.80 & - & 51.76 & 37.88 & 55.62 & - & 21.49 & - & 47.85 & - & 27.32 \\
            \hline
    		Finetune & 21.91 & 24.45 & 18.97 & 25.99 & 32.85 & 13.83 & 19.21 & 20.98 & 14.6 & 22.76 & 38.74 & 11.57 & 28.90 & 33.59 & 25.26 & 25.30 \\
    		LwF \citep{li2017learning} & 21.69 & 24.70 & 17.13 & 26.85 & 37.41 & 14.69 & 23.90 & 25.15 & 16.15 & 27.65 & 40.61 & 15.11 & 30.32 & 34.66 & 31.46 & 26.95 \\
    		% ER \cite{riemer2018learning} & 23.85 & 25.96 & 16.07 & 28.76 & 27.19 & 22.78 & 23.94 & 19.06 & 25.29 & 24.87  \\
            PODNet-flat \citep{douillard2020podnet} & 27.82 & 29.72 & 24.39 & 31.23 & 39.39 & 15.91 & 24.18 & 25.25 & 16.21 & 27.77 & 41.21 & 15.15 & 31.64 & 35.87 & 27.38 & 28.33 \\
            General-Inc \citep{xie2022general} & 31.75 & 31.63 & 23.20 & 34.01 & 44.20 & 23.16 & 34.09 & 34.20 & 24.02 & 37.04 & 48.85 & 23.40 & 35.21 & 38.82 & 30.37 & 32.23 \\
            % ESMER \cite{sarfraz2023error} & 24.65 & 24.05 & 15.78 & 26.40 & 28.99 & 26.04 & 25.64 & 20.41 & 27.10 & 31.75   \\
            PCR \citep{lin2023pcr} & 24.87 & 26.01 & 21.24 & 27.36 & 34.79 & 17.40 & 31.51 & 31.94 & 26.28 & 33.52 & 44.12 & 21.67 & 28.83 & 32.78 & 27.56 & 27.33 \\
            PRD \citep{asadi2023prototype} & 34.78 & 33.85 & 25.26 & 36.28 & 44.92 & 25.09 & 36.32 & 36.18 & 25.39 & 39.19 & 48.10 & 25.39 & 36.63 & 39.35 & 31.02 & 32.88 \\
            General-Inc+VCL \citep{xie2022general,hou2020visual} & 30.45 & 30.65 & 24.13 & 32.49 & 42.17 & 22.03 & 33.10 & 33.29 & 22.94 & 36.17 & 47.76 & 23.18 & 34.39 & 38.16 & 30.72 & 31.14 \\ 
            General-Inc+SCL \citep{xie2022general,hou2022discovering} & 31.11 & 31.28 & 23.89 & 33.37 & 43.12 & 22.65 & 34.42 & 34.56 & 23.87 & 37.54 & 48.44 & 22.74 & 34.11 & 37.88 & 29.92 & 30.09 \\
            \hline
    	IRD (Ours) & \textbf{36.18} & \textbf{34.64} & \textbf{26.86} & \textbf{36.84} & \textbf{47.49} & \textbf{26.52} & \textbf{37.45} & \textbf{37.22} & \textbf{26.66} & \textbf{40.16} & \textbf{52.55} & \textbf{26.21} & \textbf{37.69} & \textbf{41.42} & \textbf{32.87} & \textbf{33.69} \\
    		%  \hline
            \bottomrule[1pt]
    	\end{tabular}
    }
\label{tab:all-res}
% \vspace{-10pt}
\end{table*}
\subsubsection{Evaluation Metrics}
In the IHOID setup, we adopt the mean Average Precision (mAP) as the primary evaluation metric for both datasets, aligning with the standard test setting of HICO-DET.
The matching criterion for a detected human-object pair hinges on the intersection over union (IoU) between the predicted and ground truth bounding boxes for both human and object. A pair is deemed a match if the IoU surpasses 0.5. Among these matched pairs, the one with the highest score is labeled as a true positive, while others are regarded as false positives. Any pair lacking a corresponding ground truth match is also classified as a false positive.
% The matching criterion for a detected human-object pair is based on the minimum intersection over union (IoU) between the human and object bounding boxes and their corresponding ground truth counterpart. If the IoU exceeds 0.5, then the pair is considered a match with a ground truth pair. Among all the matched pairs, the one with the highest score is considered the true positive, while the remaining matched pairs are considered false positives. Pairs that do not have a matched ground truth are also considered false positives.

To evaluate the model's performance on all learned HOI categories, we test the mAP of new HOI categories and all old HOI categories (\textit{Old} in Tab.~\ref{tab:all-res}) by the end of each time phase. The combination of these two parts is denoted as \textit{Full}. We also evaluate two other category sets within HICO-DET by following the setup in \citet{chao:wacv2018}: HOI categories with less than 10 training instances (\textit{Rare}) and the remaining ones (\textit{Non-rare}).
To evaluate the model's resilience to Interaction Drift (\textit{RID}), we conduct tests on a subset of HOI categories that include relation classes appearing in both the current and previous phases, and we elaborate on the calculation scheme of RID in Suppl.
Additionally, the generalization performance on zero-shot HOIs is demonstrated by testing models on unseen HOI combinations (\textit{UC}) until the end of each learning phase.

% \vspace{-15pt}
\subsubsection{Implementation Details}
% Regarding the object detector, we adopt the H-Deformable DETR model pretrained on the HICO-DET and V-COCO datasets before incremental training, subsequently freezing its weights. All experiments on each dataset use the same DETR weight for a fair comparison.
% Specifically for HICO-DET, we utilize publicly available DETR models that are pre-trained on MS COCO \cite{lin2014microsoft}.
% However, for V-COCO, to avoid overlap with its test set present in the COCO val2017 subset, we train DETR models from scratch on MS COCO, excluding images from the V-COCO test set. Across all setups, ResNet-50 \cite{He2015} serves as our backbone for image feature extraction.
% \textcolor{red}{Regarding the object detector, following PViC \cite{zhang2023exploring}, we adopt the H-Deformable DETR model pretrained on the HICO-DET and V-COCO datasets for subsequent experiments on these two datasets, respectively. All experiments on each individual dataset use the same detector weight for fair comparisons.}
% As for the input to the relation branch, the post-processing of detection results obtained from \textcolor{red}{H-Deformable DETR follows the same approach as PViC \cite{zhang2023exploring}}. 

For the object detector, we use the H-Deformable DETR (Swin Large) model \citep{jia2023detrs} pretrained on HICO-DET and V-COCO datasets respectively, following the methodology outlined in PViC. All experiments on each dataset use the same detector weight for fair comparisons. The post-processing of detection results for the relation branch input follows the procedure detailed in PViC.
The capacity of each queue in the concept-feature dictionary $L$ is $10$, the final scores exponential parameter $\lambda$ is $0.26$ \citep{zhang2023exploring}, and the momentum value $m$ is set as $0.999$ \citep{he2020momentum}.
During training, we utilize the AdamW optimizer with a total of 25 epochs for each learning phase. The learning rate is initially set at $10^{-4}$ and decreases by a factor of 10 after the 17th epoch is finished. The coefficients of the loss terms are set as $\alpha_0=2.5, \alpha_1=0.05, \alpha_2=0.05$. The training is conducted on 8 GPUs, with a batch size of 8 per GPU.

% \vspace{-10pt}
\subsection{Results}
\label{sec:exp_res}

Here we summarize the experimental results for the IHOID setting on both the HICO-DET and V-COCO datasets. The tables show the results after the last training phase. Tab.~\ref{tab:all-res} demonstrates that our IRD consistently outperforms the baselines in alleviating forgetting, resolving interaction drift, and generalizing to zero-shot combinations on both datasets.

% \vspace{-15pt}
\subsubsection{Catastrophic Forgetting}
Our model effectively mitigates forgetting, achieving mAP of 36.18\% and 37.45\% on HICO-DET old classes,
% , and shows better stability-plasticity balance on full categories with mAP of 34.64\% and 37.22\% on HICO-DET with 5 phases and 10 phases, respectively. On V-COCO, our method also achieves SOTA performance with about 2.6\% mAP improvement compared to baselines.
and shows a better stability-plasticity balance with mAPs of 34.64\% and 37.22\% on HICO-DET for 5 and 10 phases, respectively. On V-COCO, it surpasses PRD with a 2.2\% mAP increase, marking state-of-the-art performance.
% PODNet-flat performs better in alleviating forgetting compared to other baselines because it belongs to the category of feature distillation methods.

% \vspace{-15pt}
\subsubsection{Robustness to Interaction Drift}
% Our model demonstrates advantages in addressing the challenges of interaction drift as reflected in the metrics of Robustness to Interaction Drift (RID). 
On HICO-DET, our model surpasses the best baseline by more than 2.5\% and 4.4\% under 5-phase and 10-phase setups, respectively, and achieves over 32\% mAP on V-COCO. This is partly attributed to the better knowledge retention of old concepts by our model. Additionally, our model learns invariant relation representations for samples with the same relation but different HOI classes, enabling generalization to new object-relation pairs while preventing the drift of old categories.
% The visualization of representations further supports this. Furthermore, the performance on Rare HOIs demonstrates our model's consideration for less frequent categories.

% \begin{figure}[!t]
% \vspace{-10pt}
%     \centering
%     \begin{subfigure}{0.49\linewidth}
% 		\centering
% 		\includegraphics[width=0.6\linewidth]{imgs/step0-v2.png}
% 		\caption{Learning phase 1}
% 		\label{fig:step0}
% 	\end{subfigure}
% 	\hfill
% 	\begin{subfigure}{0.49\linewidth}
% 		\centering
% 		\includegraphics[width=0.6\linewidth]{imgs/step4-v2.png}
% 		\caption{Learning phase 5}
% 		\label{fig:step1}
% 	\end{subfigure}
%     \caption{The incremental HOI detection visualization of the IRD model. It learns \texttt{eat pizza} at learning phase 1, then continually learns \texttt{sit\_on bench} while preserving knowledge of the previously learned HOI category.}
%     \label{fig:detect-vis}
% \vspace{-15pt}
% \end{figure}
\begin{figure}[t]
    \centering
    \subfloat[Performance on HICO-DET within 5 learning phases]{
        \centering
        \includegraphics[width=\linewidth]{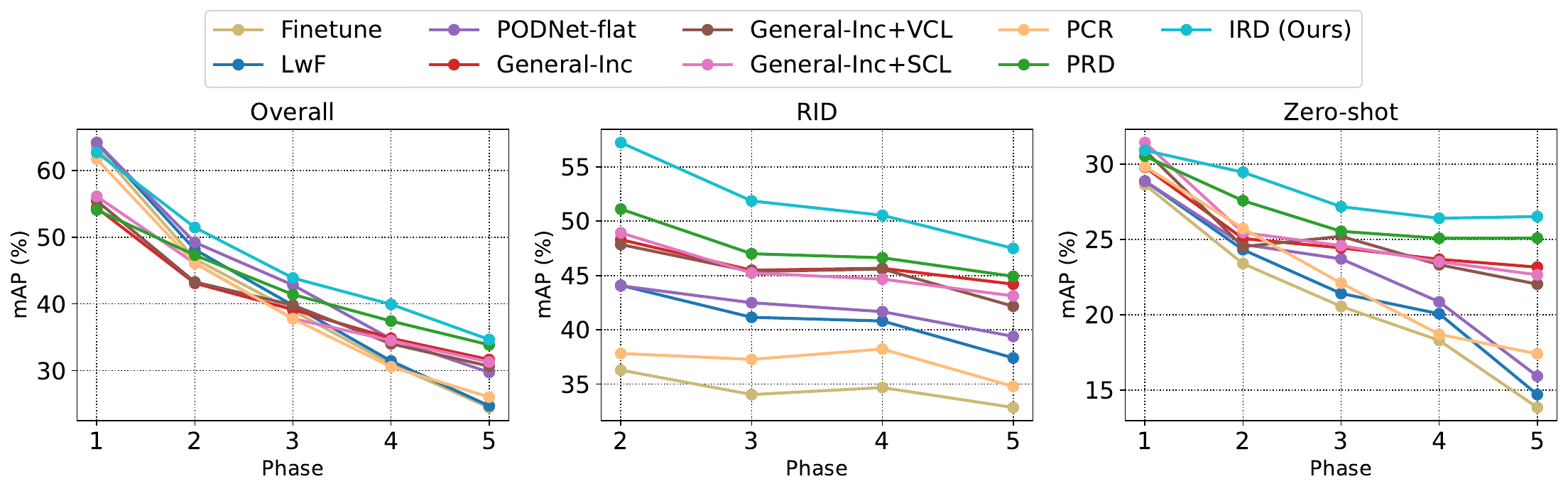}
        \label{fig:plot-hico5}
    }
    \hfill
    \subfloat[Performance on HICO-DET within 10 learning phases]{
        \centering
        \includegraphics[width=\linewidth]{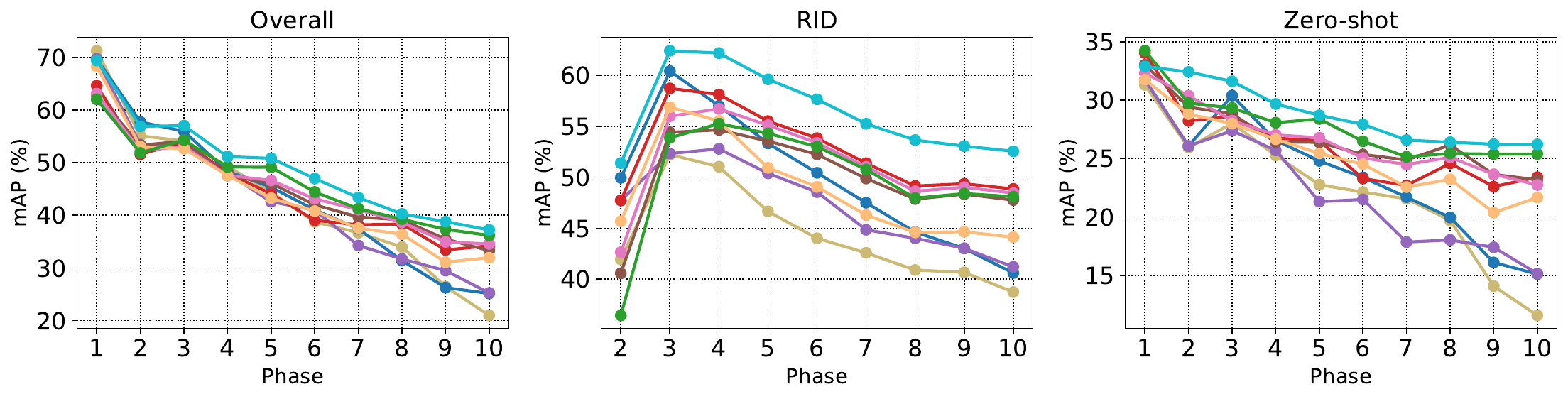}
        \label{fig:plot-hico10}
    }
    \hfill
    \subfloat[Performance on V-COCO within 5 learning phases]{
        \centering
        \includegraphics[width=\linewidth]{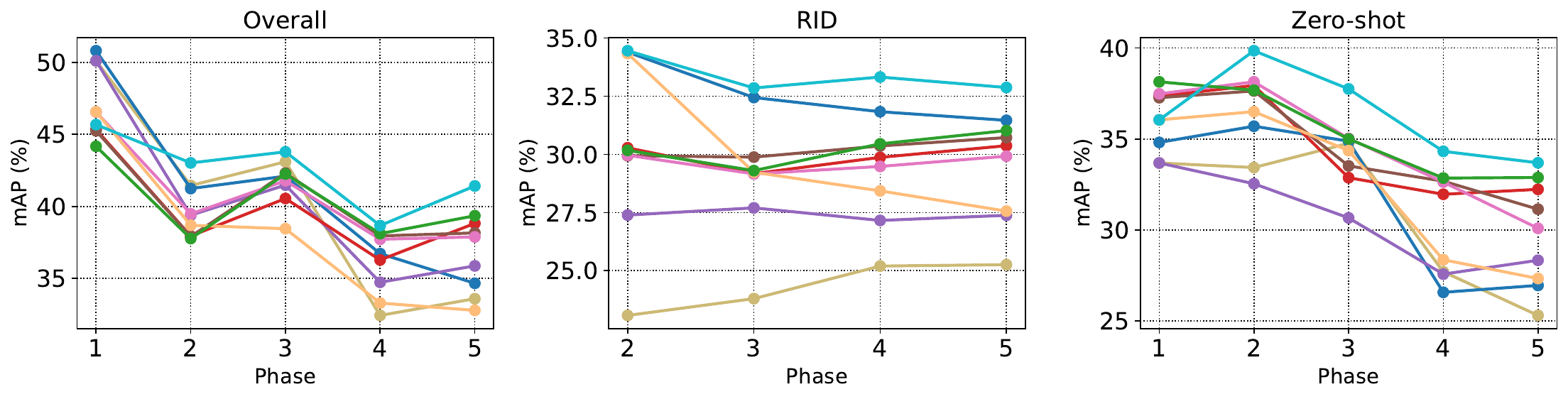}
        \label{fig:plot-vcoco5}
    }
    % \vspace{-10pt}
    \caption{Performances w.r.t. learning phases on HICO-DET and V-COCO benchmarks for overall performance (Overall), robustness to interaction drift (RID), and zero-shot detection performance (Zero-shot).}
    \label{fig:plots}
    % \vspace{-10pt}
\end{figure}

\begin{figure}[!t]
    \centering
    \subfloat[Finetune]{
        \centering
        \includegraphics[width=0.3\linewidth]{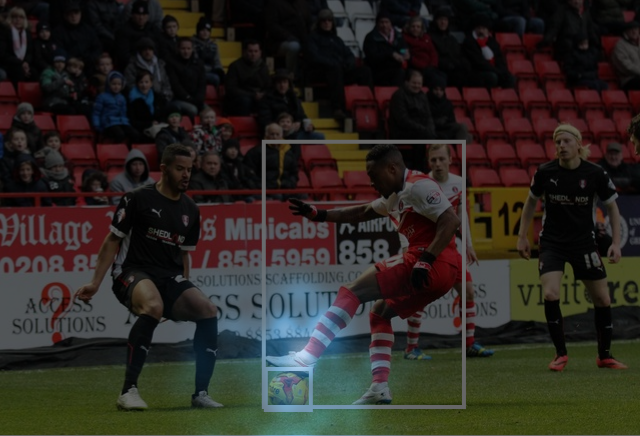}
        \label{fig:ft-step0}
    }
    \hfill
    \subfloat[General-Inc]{
        \centering
        \includegraphics[width=0.3\linewidth]{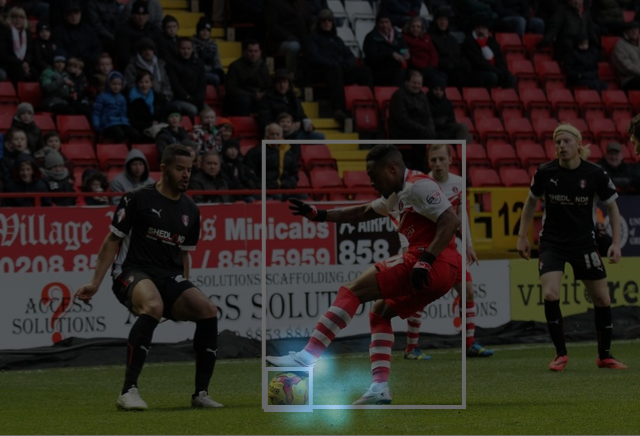}
        \label{fig:generinc-step0}
    }
    \hfill
    \subfloat[IRD (Ours)]{
        \centering
        \includegraphics[width=0.3\linewidth]{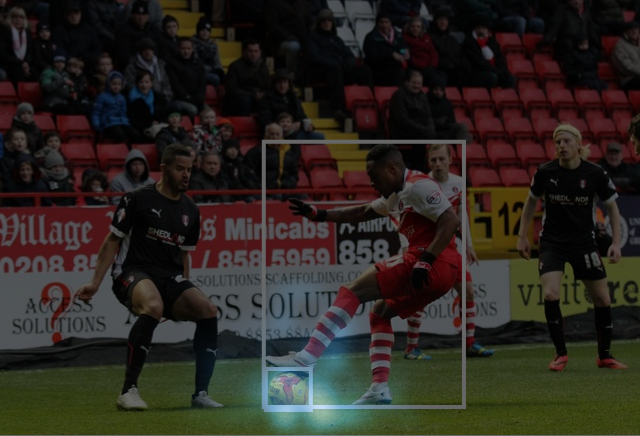}
        \label{fig:ours-step0}
    }
    % \hfill
    % \subfloat[IRD (Ours)]{
    %     \centering
    %     \includegraphics[width=0.2\linewidth]{imgs/step0-v2.png}
    %     \label{fig:det-step0}
    % }
    \\
    \subfloat[Finetune]{
        \centering
        \includegraphics[width=0.3\linewidth]{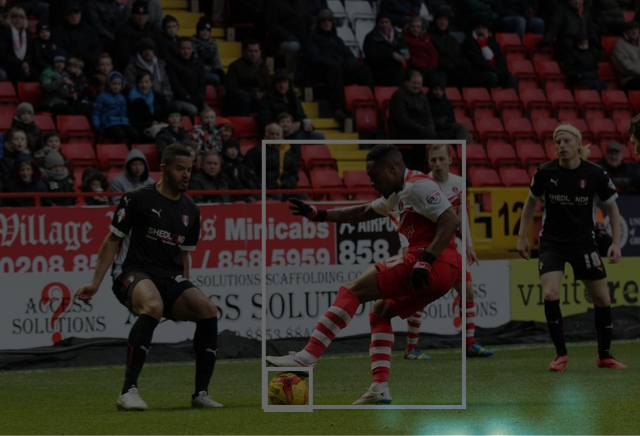}
        \label{fig:ft-step4}
    }
    \hfill
    \subfloat[General-Inc]{
        \centering
        \includegraphics[width=0.3\linewidth]{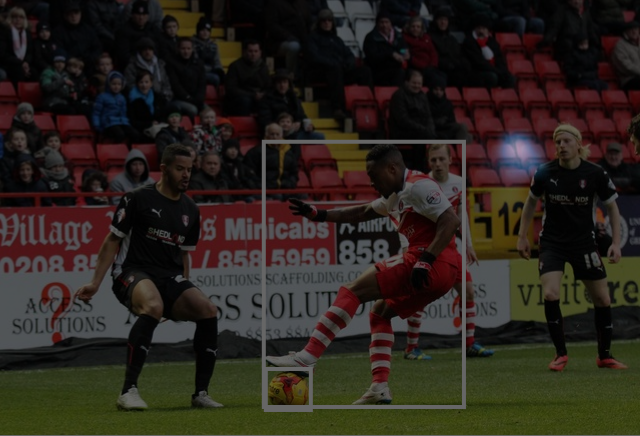}
        \label{fig:generinc-step4}
    }
    \hfill
    \subfloat[IRD (Ours)]{
        \centering
        \includegraphics[width=0.3\linewidth]{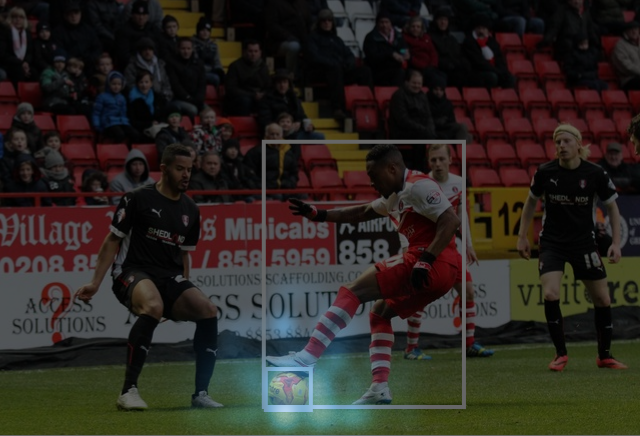}
        \label{fig:ours-step4}
    }
    % \hfill
    % \subfloat[IRD (Ours)]{
    %     \centering
    %     \includegraphics[width=0.2\linewidth]{imgs/step4-v2.png}
    %     \label{fig:det-step4}
    % }
    % \vspace{-5pt}
    \caption{
    % The comparison of the visualization results of baselines and IRD in the 5-phase incremental setting. \textbf{(a)-(d)} shows the evaluation after the 1st learning phase, while \textbf{(e)-(h)} correspond to the 5th learning phase.
     The comparison between the visualization results of baselines and IRD in the 5-phase incremental setting. \textbf{(a)-(c)} depict the results following the 1st learning phase, whereas \textbf{(d)-(f)} illustrate the results after completing the 5th learning phase.}
    \label{fig:vis}
\vspace{-3pt}
\end{figure}

% \vspace{-15pt}
\subsubsection{Zero-shot HOI Detection}
In zero-shot HOI detection shown by the \textit{UC} in Tab.~\ref{tab:all-res}, our model not only achieves SOTA performances on HICO-DET and V-COCO compared with baselines, respectively, but also surpasses the models trained in the joint training scenario. This is because the joint training model, unlike our CFD loss, only uses focal loss for learning and does not consider maintaining the consistency of representations among samples within the same relation class. The VCL and SCL methods almost show no improvement, partly due to the noise introduced by the unconstrained combination of object and relation features. Moreover, such data augmentation can only generate limited new combinations within a single training phase and cannot handle zero-shot combinations consisting of object and relation classes that appear in different phases.

% \vspace{-15pt}
\subsubsection{Per-phase Learning Performance}
We present curves of performance w.r.t. phases in Fig.~\ref{fig:plots}, where Fig.~\ref{fig:plot-hico5}, Fig.~\ref{fig:plot-hico10}, and Fig.~\ref{fig:plot-vcoco5} respectively show the performance on HICO-DET dataset within 5 phases, 10 phases, and V-COCO dataset within 5 phases. These demonstrate that our method maintains a consistent advantage throughout the learning process.

% \vspace{-15pt}
\subsubsection{Visualization}
We visualize the incremental learning results of our IRD model and comparison with baselines in Fig.~\ref{fig:vis}. Illustrated by Fig.~\ref{fig:ft-step0}-\ref{fig:ours-step0} and Fig.~\ref{fig:ft-step4}-\ref{fig:ours-step4}, the HOI \texttt{kick sports\_ball} is learned at learning phase 1, and the action \texttt{kick} is never learned again afterward. Compared with baselines, IRD focuses more on where the interaction occurs at learning phase 5.
\subsection{Ablation Study}
% \label{sec:exp_abl}
% \begin{table}[!t]
% 	\centering
% 	\caption{Ablation study on HICO-DET under the 5-phase setting.}
%  % \vspace{-10pt}
% 	\resizebox{0.6\textwidth}{!}{
%         \renewcommand{\arraystretch}{1.1}
% 		\begin{tabular}{ccc|cccccc}
% 		\toprule[1pt]
% 		CDD & MFD & CFD & Old  & Full  & Rare  & Non-Rare & RID & UC  \\\hline  
% 		$\checkmark$ & -  & -  & 28.99 & 30.45 & 21.21 & 33.06 & 40.05 & 20.04 \\
% 		$\checkmark$ & -  & $\checkmark$ & 32.82 & 33.08 & 25.84 & 35.13 & 40.13 & 22.97  \\
% 		$\checkmark$ & $\checkmark$  & - & 32.48 & 32.94 & 24.44 & 35.35 & 45.44 & 22.89  \\
% 		\rowcolor{gray!20} $\checkmark$ & $\checkmark$  & $\checkmark$ & \textbf{36.18} & \textbf{34.64} & \textbf{26.86} & \textbf{36.84} & \textbf{47.49} & \textbf{26.52} \\ \bottomrule[1pt]
% 	\end{tabular}
% }
% \label{tab:abl}
% \vspace{-15pt}
% \end{table}

To assess the necessity and effectiveness of our proposed two distillations in the IRD framework, ablative experiments are conducted on the HICO-DET dataset, starting with the naive model with $\mathcal{L}_{rel}$ and $\mathcal{L}_{CDD}$. The results are summarized in Tab.~\ref{tab:abl}.
The CFD component significantly improves the performance of unseen combinations and that of previously learned HOIs. It enhances the model's stability and generalization capability by maintaining invariant relation representations for samples with the same relation class but different HOI classes across different phases.
The MFD component aims to ensure learning robust relation representations, effectively mitigating the issue of forgetting.
With their unique roles, these two components thereby greatly enhance the model's capability of tackling interaction drift and zero-shot scenarios.
\label{sec:exp_abl}
\begin{table}[!t]
    \footnotesize
	\centering
	\caption{Ablation study on HICO-DET under the 5-phase setting.}
 % \vspace{-10pt}
	\resizebox{0.48\textwidth}{!}{
        \renewcommand{\arraystretch}{1.1}
		\begin{tabular}{ccc|cccccc}
		\toprule[1pt]
		CDD & MFD & CFD & Old  & Full  & Rare  & Non-Rare & RID & UC  \\\hline  
		$\checkmark$ & -  & -  & 28.99 & 30.45 & 21.21 & 33.06 & 40.05 & 20.04 \\
		$\checkmark$ & -  & $\checkmark$ & 32.82 & 33.08 & 25.84 & 35.13 & 40.13 & 22.97  \\
		$\checkmark$ & $\checkmark$  & - & 32.48 & 32.94 & 24.44 & 35.35 & 45.44 & 22.89  \\
		$\checkmark$ & $\checkmark$  & $\checkmark$ & \textbf{36.18} & \textbf{34.64} & \textbf{26.86} & \textbf{36.84} & \textbf{47.49} & \textbf{26.52} \\ \bottomrule[1pt]
	\end{tabular}
}
\label{tab:abl}
\vspace{-12pt}
\end{table}
% \begin{figure}[!t]
%     \centering
%     \includegraphics[width=\linewidth]{imgs/zero-shot.pdf}
%     \caption{Zero-shot performances w.r.t. steps on HICO-DET.}
%     \label{fig:hico-zs}
%     \vspace{-10pt}
% \end{figure}

% \vspace{-10pt}
\section{Conclusion}
% \vspace{-5pt}
In summary, we introduce the incremental learning setting for human-object interaction detection, which is accompanied by three challenges including forgetting previously learned HOI categories, the interaction drift on the relation classes that appear across multiple learning phases, and the difficulty in generalizing to zero-shot HOI combinations. Our proposed incremental relation distillation framework offers a novel approach by first disentangling the learning of objects and relations and then emphasizing the acquisition of robust and invariant relation representations through carefully designed distillations. These distillation losses are supported by a momentum teacher and a dynamically updated concept-feature dictionary. Through extensive experiments on the HICO-DET and V-COCO datasets, we have demonstrated the effectiveness of our method to tackle all three challenges.

\clearpage
\newpage

\appendix
% \clearpage
% The supplementary material is organized as follows. In Sec.~\ref{sec:sir-base}, we provide detailed information on our proposed SIR and how we apply baselines in our IHOID setup. Sec.~\ref{sec:suppl-setup} presents more about the experiment setup including dataset partitioning and evaluation metrics. In Sec.~\ref{sec:suppl-res}, we show more experiment results including model performance after each training step and more ablation studies. 
% The supplementary material is organized as follows. In Sec.~\ref{sec:sir-base}, we provide detailed information on our proposed IRD method and explain how we apply baselines in our IHOID setup. Sec.~\ref{sec:suppl-setup} provides extra details about the experiment setup, including dataset partitioning and evaluation metrics. In Sec.~\ref{sec:suppl-hicores}, we present additional experiment results on HICO-DET (5 steps), including further ablation studies and t-SNE visualizations of relation features. In Sec.~\ref{sec:suppl-vcocores}, we present per-step performances on V-COCO under the 5-step setting.

% , such as  further sensitive studies, generalizability experiments of IRD,and t-SNE visualizations of relation features
% Finally, Sec.~\ref{sec:suppl-vcocores} discusses per-phase performance on V-COCO.

% To further validate the robustness of IRD to the three challenges, we conduct experiments on a large-scale and long-tailed dataset SWiG-HOI \cite{wang2021discovering} and demonstrate the performance of IRD and comparisons to baselines in Sec.~\ref{sec:suppl-swigres}.
% \vspace{-10pt}
% \section{Appendix}
\label{sec:appendix}

The appendix is structured as follows: Sec.~\ref{sec:sir-base} adapts baseline methods in the IHOID setup. Sec.~\ref{sec:suppl-setup} elaborates on the experimental setup, including dataset partitioning and evaluation metrics. Additional results and analyses on HICO-DET are presented in Sec.~\ref{sec:suppl-hicores}, and Sec.~\ref{sec:tsne} shows the t-SNE visualization of relation features. The \textcolor{red}{red} numbers of sections and tables refer to those in the main text, while \textcolor{iccvblue}{blue} numbers refer to those in the appendix.

\section{Adaptation of Baselines in IHOID}
\label{sec:sir-base}

In this section, we detail the adaptation of the baselines mentioned in Sec.~\textcolor{red}{5.1} to our framework in the IHOID setting. This includes LwF \cite{li2017learning}, PODNet \cite{douillard2020podnet}, General-Inc \cite{xie2022general}, PCR~\cite{lin2023pcr}, PRD~\cite{asadi2023prototype}, General-Inc+VCL \cite{xie2022general,hou2020visual}, and General-Inc+SCL~\cite{xie2022general,hou2022discovering}. Given the differences in data and model structures between the CIL and IHOID tasks, we have retained the original implementations of LwF, while the other baselines were modified to fit our specific context.
% \vspace{5pt}

% \paragraph{ESMER} We retain the core weighting strategy of ESMER but modify the episodic memory maintenance. Since an image can contain multiple human-object pairs, we select images to be stored in memory based on the following criteria: the predictions of the image should include box pairs with an IoU greater than 0.5 with the ground truth pairs, and simultaneously the focal loss values for these pairs meet the requirements for low-loss samples as described in the paper~\cite{sarfraz2023error}.
% \paragraph{ESMER} While preserving ESMER's fundamental weighting strategy, we have adapted its episodic memory management. In our approach, an image is selected for memory storage if it meets two criteria: first, its predicted box pairs must have an Intersection over Union (IoU) greater than 0.5 with the ground truth pairs, and second, the focal loss values for these pairs should classify them as low-loss samples, in line with the guidelines set out in \cite{sarfraz2023error}.
% \vspace{-5pt}

\paragraph{PODNet.} We mainly adopt the feature distillation idea from PODNet. Since the whole module before the relation classifier is Transformer-based rather than CNN-based as designed in PODNet, we discard the spatial distillation loss and only retain the distillation of the final embedding, which is denoted as \textit{POD-flat} in the original paper \cite{douillard2020podnet}. Specifically, we take the box pair information into the models from both phase $t-1$ and phase $t$, and calculate $\mathcal{L}_{POD-flat}$ using the output relation representations. This modified baseline method is referred to as \textbf{PODNet-flat} in our paper.
% \vspace{-8pt}

\paragraph{General-Inc.} In the IHOID setting, the problem of interaction drift is similar to the continuous domain shift for data belonging to the same category in the general incremental learning setting \cite{xie2022general}. 
% We utilize the idea of maintaining multiple prototypes per category and dynamic prototype growth in General-Inc. Specifically, when the arriving new data related to relation class $R$ involves $n$ object categories, we expand $n$ additional prototypes for relation class $R$.
Drawing from General-Inc's strategy, we adopt the concept of maintaining and dynamically expanding multiple prototypes per category. Specifically, for each new data related to a relation class $R$ that involves $n$ object categories, we create $n$ additional prototypes for class $R$. 
% \vspace{-8pt}

\paragraph{PCR.}
% We incorporate proxy-based contrastive replay (PCR) into our framework because the designs of memory replay and contrastive-based loss in PCR fit well in our IHOID setting. We concatenate original and augmented samples as the model's input and adopt the proxy-based classifier in the training stage, and follow the same inference process mentioned in the paper \cite{lin2023pcr}.
% We integrate Proxy-based Contrastive Replay (PCR) into our framework, as outlined in \cite{lin2023pcr}, due to its compatible memory replay approach and contrastive-based loss, which align well with our IHOID setting. We utilize both original and augmented samples as inputs to the model, employing the proxy-based classifier during training. For inference, we follow the same process mentioned in the paper.
We integrate a proxy-based contrastive approach into our framework, as outlined in \cite{lin2023pcr}. Concretely, we utilize both original and augmented samples as inputs to the model, employing the proxy-based classifier during training. For inference, we follow the same process mentioned in the paper. To maintain a fair comparison within our exemplar-free IHOID framework, we adjust the memory buffer size in the original PCR to zero.
% \vspace{-8pt}

\paragraph{PRD.}
In the CIL setup, the core of PRD involves three types of loss: contrastive loss, similarity loss, and distillation losses, all aimed at generating one prototype for each image category. When adapting this approach, we apply these losses to create one prototype for each relation category. We start by extracting relation features and their ground truths from images as a basis for computing loss. Specifically, we utilize features from box pairs with an IoU greater than 0.5 with the ground truth, where the used labels match the closest ground truth pair's relation label. Based on this, we fully incorporate the design of the three PRD losses into our model architecture.

% \vspace{-8pt}

\paragraph{VCL.} VCL was originally designed for zero-shot HOI detection in the joint training setting. We adapt the idea of recombining object features and relation features from different images for data augmentation in VCL. In our IHOID setting, we recombine human box features and object box features from different images. This modified VCL method serves as a plugin in our framework. As a result, we introduce a baseline method for zero-shot HOI detection in the incremental learning setup, denoted as \textbf{General-Inc+VCL}, which merges the General-Inc approach with the modified VCL technique.
% \vspace{-8pt}

\paragraph{SCL.} SCL tackles the same problem setting as VCL. Building upon the ideas of VCL, SCL further introduces the concept confidence matrix which is essentially the cross-product space of objects and relations. This enables many more combinations than VCL so that zero-shot HOIs can be detected more effectively during inference. In each learning phase of our incremental setting, we separately maintain the confidence matrix and dynamically update the confidence scores during training. We add the \textit{concept discovery loss} term corresponding to SCL to the baseline with VCL, giving \textbf{General-Inc+SCL}, which combines the General-Inc approach with SCL.

% \vspace{-8pt}
\section{Experiment Setup}
\label{sec:suppl-setup}
% In this section, we list the detailed statistics of dataset partition under the IHOID setting in Tab.~\ref{tb:updated-stats},~\ref{tb:hico-sta} and~\ref{tb:vcoco-sta}. Specifically speaking, the first four rows display the number of HOI categories, relation categories, object categories, and images involved in each training subset, respectively. The fifth row \textbf{Drift Interaction} is denoted as all HOIs encountering interaction drift problem explained in Sec.~\textcolor{red}{1}. The last row \textbf{Unseen Combination} denotes the number of zero-shot HOI combinations.

\subsection{Statistics of Preprocessed Datasets}
\label{sec:prepro} 
Tab.~\ref{tab:stat} presents the statistics of preprocessed HICO-DET and V-COCO datasets mentioned in Sec.~\textcolor{red}{5.1} for the IHOID setting, where we exclude the HOI categories specified in Sec.~\textcolor{red}{5.1.2} of the main text from both the training and test sets. 
For HICO-DET, the original dataset comprises 37,633 training images, 9,546 test images, 80 object categories, 117 action categories, and 600 HOI categories. The following table presents the statistics after preprocessing. For VCOCO, we use the dataset as processed by \cite{zhang2022efficient}, which aligns with our requirements.
% In addition to excluding the HOI categories specified in lines 347-354 of the main text from both the training and test sets, we also remove categories involving unseen combinations from the training dataset.
\begin{table*}[!t]
    \centering
    % \caption{Statistics of HICO-DET and V-COCO (in format \textit{before/after} preprocessing).}
    \caption{Statistics of Preprocessed HICO-DET and V-COCO.}
    \resizebox{\textwidth}{!}{
        % \renewcommand{\arraystretch}{1.1}
        % \fontsize{8pt}{14pt}\selectfont
        \begin{tabular}{c|c|c|c|c|c}
            \toprule[1pt]
            Datasets & \# training images & \# test images & \# object categories & \# action categories & \# HOI categories \\
            \hline
            % HICO-DET & 37,633/33,601 & 9,546/8,528 & 80/80 & 117/116 & 600/520 \\
            % V-COCO & 5,400/3,923 & 4,946/3,501 & 80/80 & 29/24 & 287/259 \\
            HICO-DET & 33,601 & 8,528 & 80 & 116 & 520 \\
            V-COCO & 5,400 & 4,946 & 80 & 24 & 287 \\
            \bottomrule[1pt]
        \end{tabular}
    }
    \label{tab:stat}
\end{table*}

\begin{table*}[!t]
    \centering
    \renewcommand{\arraystretch}{1.1}
    \caption{Statistics of the HICO-DET and V-COCO datasets in the 5-phase setup.}
    % \vspace{-10pt}
    \label{tb:5step-sta}
    \resizebox{\textwidth}{!}{
        \begin{tabular}{c|ccccc|ccccc}
            \toprule[1pt]
            & \multicolumn{5}{c|}{\textbf{HICO-DET}} & \multicolumn{5}{c}{\textbf{V-COCO}} \\
            & phase 1 & phase 2 & phase 3 & phase 4  & phase 5 & phase 1 & phase 2 & phase 3 & phase 4  & phase 5  \\ \hline
            HOI & 40    & 40    & 40    & 40     & 35  & 20	&20	&20	&20	&16   \\
            Relation & 26    & 28    & 32    & 33     & 29  & 10	&8	&7	&7	&10   \\
            Object & 30    & 32    & 29    & 27     & 24 &17	&19	&19	&15	&10    \\
            Training images & 5745  & 6178  & 2580  & 4348   & 3804 & 1088	&743	&1055	&1021	&1756  \\
            Drift Interaction & - & 16 & 26 & 34 & 30 & -     & 10 & 29 & 45 & 48 \\
            Unseen Combination & 89    & 211   & 294   & 325    & 325 & 33	&75	&118	&138	&138   \\ 
            \bottomrule[1pt]
        \end{tabular}
    }
\end{table*}

\begin{table*}[!t]
    \centering
    \footnotesize
    \renewcommand{\arraystretch}{1.1}
    \caption{Statistics of the HICO-DET dataset in the 10-phase setup.}
    % \vspace{-10pt}
    \label{tb:updated-stats}
    \resizebox{\textwidth}{!}{
        \begin{tabular}{c|cccccccccc}
            \toprule[1pt]
            & phase 1 & phase 2 & phase 3 & phase 4 & phase 5 & phase 6 & phase 7 & phase 8 & phase 9 & phase 10 \\ \hline
            HOI                                              & 17 & 17 & 17 & 17 & 17 & 17 & 17 & 17 & 17 & 17 \\
            Relation                                         & 13 & 13 & 13 & 15 & 17 & 13 & 15 & 15 & 15 & 15 \\
            Object                                           & 15 & 14 & 16 & 14 & 16 & 16 & 15 & 15 & 15 & 14 \\
            Training Images                                  & 3497 & 1837 & 1411 & 2538 & 2590 & 1496 & 1668 & 1949 & 2941 & 1203 \\
            % ID Portion                                       & - & 5/17 & 5/34 & 9/51 & 18/68 & 13/85 & 12/102 & 7/119 & 19/136 & 15/153   \\
            Drift Interaction                                       & - & 5 & 5 & 9 & 18 & 13 & 12 & 7 & 19 & 15   \\
            Unseen Combination                                               & 37 & 69 & 141 & 185 & 245 & 287 & 319 & 332 & 340 & 342  \\
            \bottomrule[1pt]
        \end{tabular}
    }
\end{table*}
% \vspace{-20pt}
\subsection{Statistics on Training Set Partition}
\label{sec:sta}
In this section, detailed statistics of dataset partitioning under the IHOID setting are presented in Tab.~\ref{tb:5step-sta} and Tab.~\ref{tb:updated-stats}. Specifically, the first four rows of each table indicate the quantities of HOI categories, relation categories, object categories, and training images, respectively. The fifth row, labeled \textbf{Drift Interaction}, represents all HOIs learned previously which are affected by the interaction drift issue discussed in lines 202-213. The final row, \textbf{Unseen Combination}, quantifies the zero-shot HOI combinations.
%当我们划分数据集时，每一步我们随机从preprocess好的数据集中取出一个子集，并满足requirements in lines 358-361 of Section~\textcolor{red}{4.1}.
When partitioning the dataset, at each phase, we randomly extract a subset from the preprocessed dataset, ensuring it meets the requirements outlined in Sec.~\textcolor{red}{5.1.3}.

% \vspace{-10pt}

% \paragraph{HICO-DET}Tab.~\ref{tb:updated-stats} and ~\ref{tb:hico-sta} present the detailed statistics of the training subset split into 10 and 5 subsets at each time step on the HICO-DET~\cite{chao:wacv2018} dataset, as mentioned in Sec.~\textcolor{red}{4.1}.
% Noticeably, At the end of steps 4 and 5, the model faces the same count of zero-shot combinations, because although step 5 introduces previously unseen relations and objects, they do not contribute valid unseen combinations, resulting in no increase in the count of zero-shot HOI combinations.

\paragraph{HICO-DET.} Tab.~\ref{tb:5step-sta} and Tab.~\ref{tb:updated-stats} detail the division of the training subset into 5 and 10 phases, respectively, for the HICO-DET dataset~\cite{chao:wacv2018}. Notably, at the end of both phases 4 and 5, the model encounters an identical number of zero-shot combinations. This is because the new relations and objects introduced in phase 5 do not form any additional valid unseen combinations, leaving the count of zero-shot HOI combinations unchanged. 
%值得注意的是，我们需要定义好inference的zero-shot HOIs，并从训练集的annotation中去除这些类别。由于分割数据集需满足line 358-361 of Sec.~4.1，且完全是随机划分的，因此在我们的5-phase和10-phase的设定下，需要inference的zero-shot HOI是不同的，因此这两个设定的训练数据不同，也就在Table1中两个设定的的upper bound性能是不同的。

Additionally, it is important to note that we need to clearly define the unseen HOI combinations for inference and remove these categories from the training set annotations. Given that the dataset split must comply with requirements in Sec.~\textcolor{red}{5.1.3}, and is entirely randomized, the unseen HOI combinations required for inference differ between our 5-phase and 10-phase setups. Consequently, the training data varies between these two settings, leading to different upper-bound performances for each setup in Tab.~\textcolor{red}{1}.
% \vspace{-10pt}

\paragraph{V-COCO.} For the V-COCO~\cite{gupta2015visual} dataset, we also follow the data partitioning described Sec.~\textcolor{red}{5.1}, and the specific statistics of subsets are presented in Tab.~\ref{tb:5step-sta}. 
V-COCO is only split into 5-phase subsets, as a 10-phase division results in too small subsets for effective training. 

% \begin{table}[t]
%     \centering
%     \renewcommand{\arraystretch}{1.1}
%     \caption{Statistics of the V-COCO dataset partitioned into 5 steps.}
%     \label{tb:vcoco-sta}
%     \resizebox{0.45\textwidth}{!}{
%         \begin{tabular}{c|ccccc}
%             \toprule[1pt]
%                                                         & step 1 & step 2 & step 3 & step 4  & step 5  \\ \hline
%             HOI                                           & 20	&20	&20	&20	&16     \\
%             Relation                                      & 10	&8	&7	&7	&10    \\
%             Object                                           &17	&19	&19	&15	&10     \\
%             Training images                                  & 1088	&743	&1055	&1021	&1756   \\
%             % ID portion                                 & -     & 10/20 & 29/40 & 45/60 & 48/80 \\
%             Drift Interaction                         & -     & 10 & 29 & 45 & 48 \\
%             Unseen Combination                                        & 33	&75	&118	&138	&138   \\
%             \bottomrule[1pt]
%         \end{tabular}
%     }
% \vspace{-5pt}
% \end{table}

\subsection{Evaluation Metrics}
As mentioned in Sec.~\textcolor{red}{5.1}, we mainly evaluate our method using three metrics: overall (\textit{Full}), robustness against interaction drift (\textit{RID}), and performance on zero-shot HOI categories (\textit{UC}), which are tested on different HOI category subsets. Here, we provide a detailed explanation of how these metrics are conducted after each training phase $t$.
% \vspace{-5pt}

\paragraph{Overall Performance.}
% First, 
For the overall performance, we measure the mAP on all the HOI categories $\mathcal{C}_{1:t}$ that have been learned up to phase $t$.
% \vspace{-12pt}

\paragraph{Robustness against Interaction Drift.}
% Second, 
For RID, we first evaluate the model's mAP on a subset $\mathcal{C}_t^{rid}$, comprising previously learned classes affected by interaction drift at each phase. Specifically, $\mathcal{C}_t^{rid}$ consists of HOI categories $C_i=(O_j, R_k)$ where $C_i\in\mathcal{C}_{1:t-1}$, $R_k\in\mathcal{R}_{1:t-1}$, and $R_k\in \mathcal{R}_{t}$ at the same time. In other words, for each old class $C_i$, the corresponding relation category has appeared in both the previous phases and the current phase. Then, we calculate the average mAP across all phases encountered, as the final numerical result presented in Tab.~\textcolor{red}{1}.
% \vspace{-12pt}

% \paragraph{Robustness against Interaction Drift}
% Secondly, to assess RID, we measure the model's mAP on a subset, $\mathcal{C}_t^{id}$, comprising previously learned classes affected by interaction drift. Specifically, $\mathcal{C}_t^{id}$ includes HOI categories $C_i=(O_j, R_k)$ where $C_i$ belongs to $\mathcal{C}_{1:t-1}$, and the relation category $R_k$ is present in both previous steps $\mathcal{R}_{1:t-1}$ and the current step $\mathcal{R}_{t}$. This means that for each previous class $C_i$, its associated relation category is recurring, appearing in both earlier and current steps.

\paragraph{Zero-shot HOI Detection.}
% Finally, 
For zero-shot HOI detection, 
We evaluate the model on a set of HOI categories $\mathcal{C}_t^{uc}$ that the model has not seen before, but they are reasonable combinations of object and relation categories based on the objects and relations the model has encountered up to the current phase. Specifically, $\mathcal{C}_t^{uc}$ consists of HOI categories $C_i=(O_j, R_k)$ where $O_j \in \mathcal{O}_{1:t}$, $R_k \in \mathcal{R}_{1:t}$, and $C_i \notin \mathcal{C}_{1:t}$.

% Lastly, for UC, we assess the model using the set $\mathcal{C}_t^{uc}$, comprising HOI categories unseen by the model yet logically inferred from known objects and relations up to the current step. Specifically, $\mathcal{C}_t^{uc}$ contains HOI categories $C_i=(O_j, R_k)$, where $O_j$ is within $\mathcal{O}_{1:t}$, $R_k$ is within $\mathcal{R}_{1:t}$, but $C_i$ itself is not part of $\mathcal{C}_{1:t}$.

% \vspace{-3pt}
\section{More Experiment Results}
\label{sec:suppl-hicores}

% As described in Sec.~\textcolor{red}{4.2}, we compare with baselines LwF \cite{li2017learning}, ER \cite{riemer2018learning}, PODNet \cite{douillard2020podnet}, and ESMER \cite{sarfraz2023error}, General-Inc \cite{xie2022general}, and PODNet +VCL \cite{douillard2020podnet,hou2020visual}.

% It is worth noting that when the baseline models perform inference on the same zero-shot cases (as mentioned in Sec.~\ref{sec:sta}) in step 5 as in step 4, their performance is lower, which is caused by forgetting previously learned relations. In contrast, our model is able to improve its zero-shot performance in step 5. This suggests that our SIR learns stable invariant representations that generalize across different samples containing the same relation class, even at different learning steps. 
% , and the learning in step 4 reinforces the invariant characteristics of the learned relation representations.

\subsection{Comparison with More Baselines}
In addition to the experiments presented in Tab.~\textcolor{red}{1}, on HICO-DET with 5 training phases, we have included two more baselines for tackling zero-shot HOIs in the incremental learning setup, as shown in Tab.~\ref{tb:prd-series}. In the main text, we combined the baseline capable of addressing the general incremental setting, specifically General-Inc with VCL and SCL. Here, we also incorporate combinations of SOTA in CIL settings, PRD, with VCL and SCL, applying both PRD's and VCL/SCL's losses to the HOI detector. The integration of VCL and SCL with PRD yields limited gains, for reasons similar to those discussed for General-Inc+VCL/SCL in lines 533-540 of Sec.~\textcolor{red}{5.2}. Our method still demonstrates superior performance on all metrics.

\begin{table}[!t]
    % \footnotesize
    \centering
    \renewcommand{\arraystretch}{1.1}
    \caption{Comparison with other baselines PRD+VCL and PRD+SCL on the HICO-DET dataset with 5 training phases.}
    % \vspace{-8pt}
    \label{tb:prd-series}
    % \fontsize{8pt}{14pt}\selectfont
    \resizebox{0.48\textwidth}{!}{
        \begin{tabular}{c|cccccc}
            \toprule[1pt]
            Methods                                        & Old	&Full	&Rare	&Non-Rare	&RID &UC \\ \hline
            % 2                                           & 	&	&	&	&  &  \\
            PRD~\cite{asadi2023prototype}                                             &34.78	&33.85	&25.26	&36.28	&44.92	&25.09 \\
            PRD+VCL~\cite{asadi2023prototype, hou2020visual}                                         &34.81	&33.90	&25.41	&36.30	&45.00	&25.36 \\
            PRD+SCL~\cite{asadi2023prototype, hou2022discovering}                                         &34.75	&33.82	&25.40	&36.20	&44.83	&25.14 \\
            IRD (Ours)                   &\textbf{36.18}	&\textbf{34.64}	&\textbf{26.86}	&\textbf{36.84}	&\textbf{47.49}	&\textbf{26.52} \\
            
            \bottomrule[1pt]
        \end{tabular}
    }
\end{table}

\subsection{More Analysis on Hyperparameters}
\label{sec:suppl-abl}
% \paragraph{Memory Size} 
% We explore the impact of memory size on our proposed method. Tab.~\ref{tb:ab-mem} presents the results of different memory sizes used in our method. It can be observed that as the memory size increases, the model's performance generally improves. Furthermore, even without using memory, our method still outperforms the baselines, indicating the superiority of our proposed method in handling the IHOID setting.
% \begin{table}[!b]
% \vspace{-5pt}
%     \centering
%     \renewcommand{\arraystretch}{1.1}
%     \caption{Analysis of the memory size.}
%     \label{tb:ab-mem}
%     \resizebox{0.45\textwidth}{!}{
%         \begin{tabular}{c|cccccc}
%             \toprule[1pt]
%             size                                            & Old	&Full	&Rare	&Non-Rare	&UC &RID \\ \hline
%             0                                           & 29.90	&30.55	&23.78	&32.46	&21.27 &31.11    \\
%             20                                      &30.04	&30.64	&23.74	&32.60     &22.56  &33.06    \\
%             50                                           & 30.69	&31.13	&23.00	&\textbf{33.40}	&22.11 &\textbf{34.46}    \\
%             100                                  & \textbf{30.96}	&\textbf{31.47}	&\textbf{24.90}	&33.32	&\textbf{22.70} &33.64  \\
%             % 200 & 29.71	&30.25	&20.72	&32.95	&22.29 \\
%             \bottomrule[1pt]
%         \end{tabular}
%     }
% \vspace{-5pt}
% \end{table}

\paragraph{Capacity of Queue.} 
In Tab.~\ref{tb:ab-L}, we show the sensitive study on the capacity $L$ of each queue in our concept-feature dictionary on the HICO-DET dataset with 10 training phases. We observe our method works better when $L=10$. 
The maximum performance difference is only 0.97\% when using different values for $L$, which indicates our method is robust to this hyperparameter. 

\begin{table}[!t]
% \vspace{-10pt}
\footnotesize
    \centering
    \renewcommand{\arraystretch}{1.1}
    \caption{\small{Sensitive study on the capacity $L$ of each queue in the concept-feature dictionary.}}
    % \vspace{-10pt}
    \label{tb:ab-L}
    % \fontsize{8pt}{14pt}\selectfont
    \resizebox{0.48\textwidth}{!}{
        \begin{tabular}{c|cccccc}
            \toprule[1pt]
            Setting                                        & Old	&Full	&Rare	&Non-Rare	&RID &UC \\ \hline
            % 2                                           & 	&	&	&	&  &  \\
            $L=5$                                              & 37.34	&37.17	&26.44	&40.16	&51.58	&25.72 \\
            $L=10$                           & \textbf{37.45}	&\textbf{37.22}	&26.66	&\textbf{40.16}	&\textbf{52.55}	&26.21   \\
            $L=20$                                             & 37.36	&37.21	&\textbf{27.17}	&40.01	&51.44  &\textbf{26.35} \\
            
            \bottomrule[1pt]
        \end{tabular}
    }
% \vspace{-15pt}
\end{table}

% \begin{figure}[!t]
%     \centering
%     \includegraphics[width=0.9\linewidth]{imgs/zero-shot_VCOCO.pdf}
%     \caption{Zero-shot performances w.r.t. steps on V-COCO under the 5-step setting.}
%     \label{fig:zs-vcoco}
% \vspace{-5pt}

\subsection{Generalizability of Our IRD}
\label{sec:upt}
\begin{figure}[!t]
    \centering
    \includegraphics[width=\linewidth]{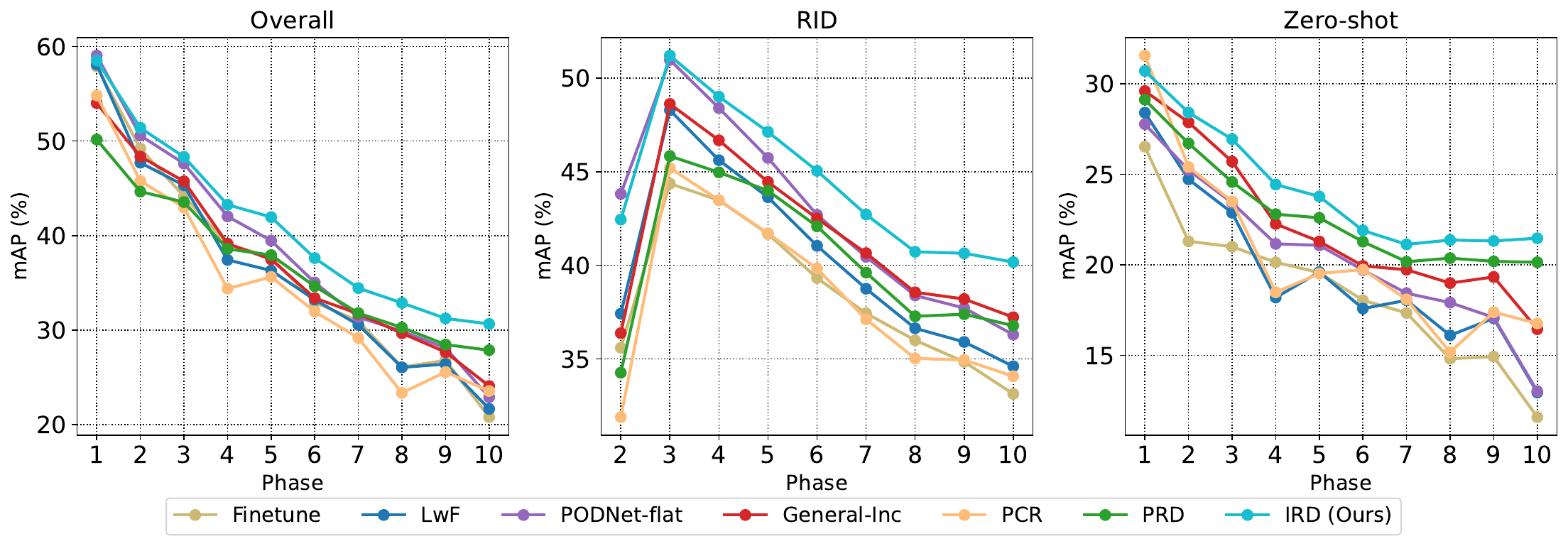}
    % \vspace{-5pt}
    \caption{Performances w.r.t. learning phases on HICO-DET benchmark under the 10-phase setting, with UPT as the basic HOI detector.}
    \label{fig:upt-hico10}
\end{figure}

To validate the generalizability of our strategy in the IHOID setup, we adopt another two-stage HOI detector UPT \cite{zhang2022efficient} as the base model and conduct additional incremental learning experiments, employing ResNet-50~\cite{He2015} as the backbone. We compare our IRD with the classic and SOTA baselines of incremental learning on the HICO-DET dataset under the 10-phase setting. As shown in Tab.~\ref{tab:upt} and Fig.~\ref{fig:upt-hico10}, our IRD method still consistently achieves the best performance on the overall, RID, and zero-shot HOI evaluation metrics. 
% 表1整体上展现出比表S6更好的性能，因为表1使用PViC的作为base HOI detector，use the Swin-L as the backbone,有着更好的基础性能。
Additionally, Tab.~\textcolor{red}{1} shows overall better performance compared to Tab.~\ref{tab:upt} due to employing PViC~\cite{zhang2023exploring} as the base HOI detector and using Swin-L~\cite{liu2021swin} as the backbone, resulting in enhanced foundational performance.

% \vspace{-15pt}
\begin{table}[!b]
\centering
\caption{Experiment results on HICO-DET dataset within 10 training phases, with UPT as the basic HOI detector.}
 % \vspace{-5pt}
    \resizebox{0.48\textwidth}{!}{
        \renewcommand{\arraystretch}{1.1}
        % \fontsize{8pt}{14pt}\selectfont
    	\begin{tabular}{c|cccccc}
    		\toprule[1pt]
    		\textbf{Methods} & Old & Full & Rare & Non-rare & RID & UC  \\ \hline
    		Joint (Upper Bound) & - & 40.06 & 30.26 & 42.78 & - & 20.44 \\
            \hline
    		Finetune & 19.22 & 20.79 & 17.38 & 21.74 & 33.14 & 11.58  \\
    		LwF \cite{li2017learning} & 20.30 & 21.66 & 15.83 & 23.28 & 34.61 & 12.94  \\
            PODNet-flat \cite{douillard2020podnet} & 21.55 & 22.9 & 18.97 & 24.0 & 36.30 & 13.01  \\
            General-Inc \cite{xie2022general} & 23.37 & 24.08 & 17.63 & 25.88 & 37.23 & 16.44 \\
            PCR \cite{lin2023pcr} & 23.05 & 23.55 & 20.67 & 24.35 & 34.08 & 16.76  \\
            PRD \cite{asadi2023prototype} & 27.80 & 27.87 & 22.52 & 29.36 & 36.78 & 20.14 \\
            \hline
    	 IRD (Ours) & \textbf{30.72} & \textbf{30.65} & \textbf{24.03} & \textbf{32.50} & \textbf{40.17} & \textbf{21.47} \\
    		%  \hline
            \bottomrule[1pt]
    	\end{tabular}
    }
\label{tab:upt}
% \vspace{-15pt}
\end{table}

\section{t-SNE Visualization}\label{sec:tsne}
% We employed the t-SNE~\cite{van2008visualizing} visualization technique to validate the robust and invariant relation features learned by our approach.
% Fig.~\ref{fig:tsne} illustrates the t-SNE visualization of the relation features extracted from the test set at the final step. The features with the same color represent those associated with the same relation category.
% Our method enables a more compact distribution of features for each relation. This observation indicates that samples with different HOI classes but the same relation class possess relation features that are invariant to the combined object classes, providing evidence of the effectiveness of our approach.

\begin{figure*}[htbp]
    \centering
    \subfloat[Finetune]{
        \centering
        \includegraphics[width=0.31\linewidth]{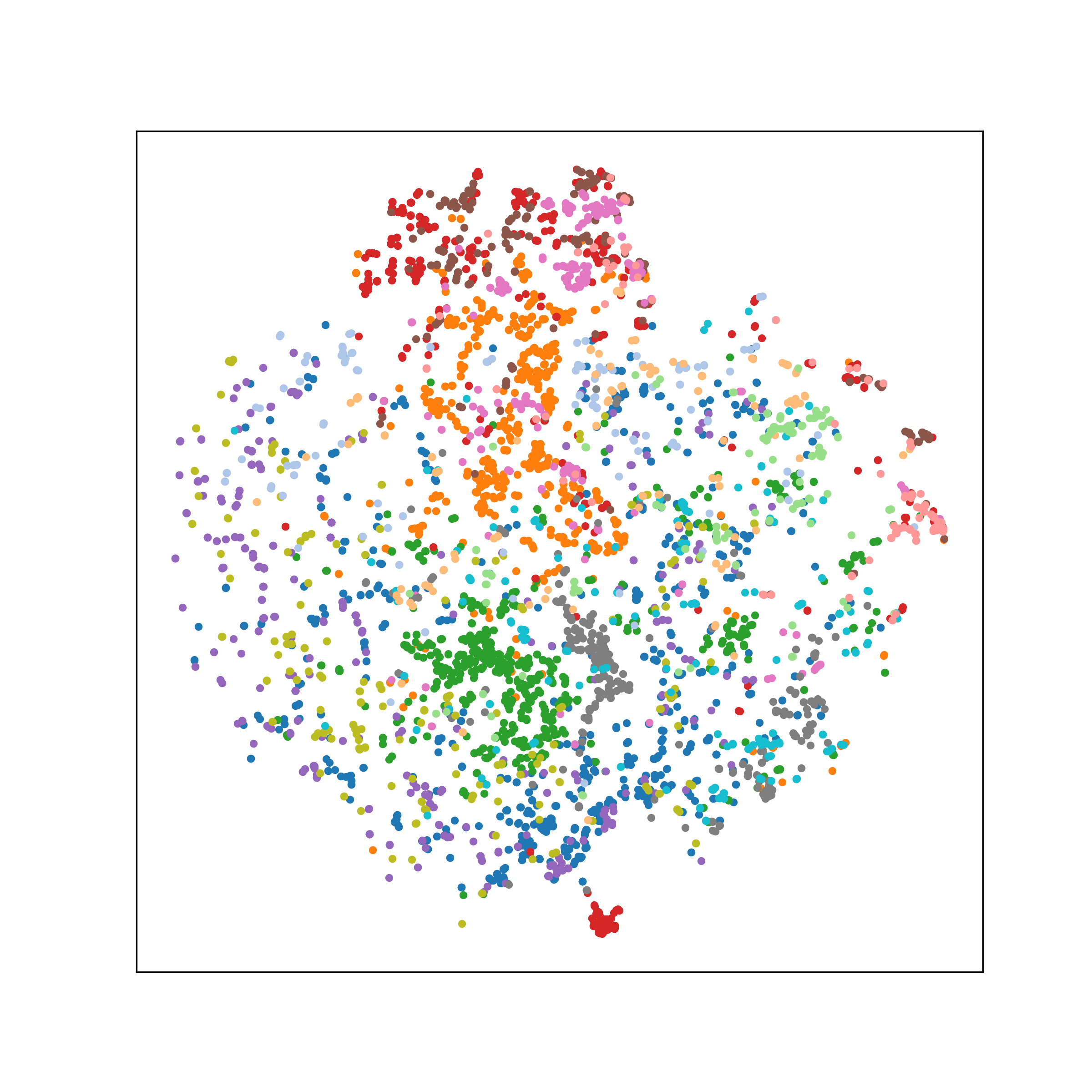}
        \label{fig:ft}
    }
    \hfill
    \subfloat[LwF]{
        \centering
        \includegraphics[width=0.31\linewidth]{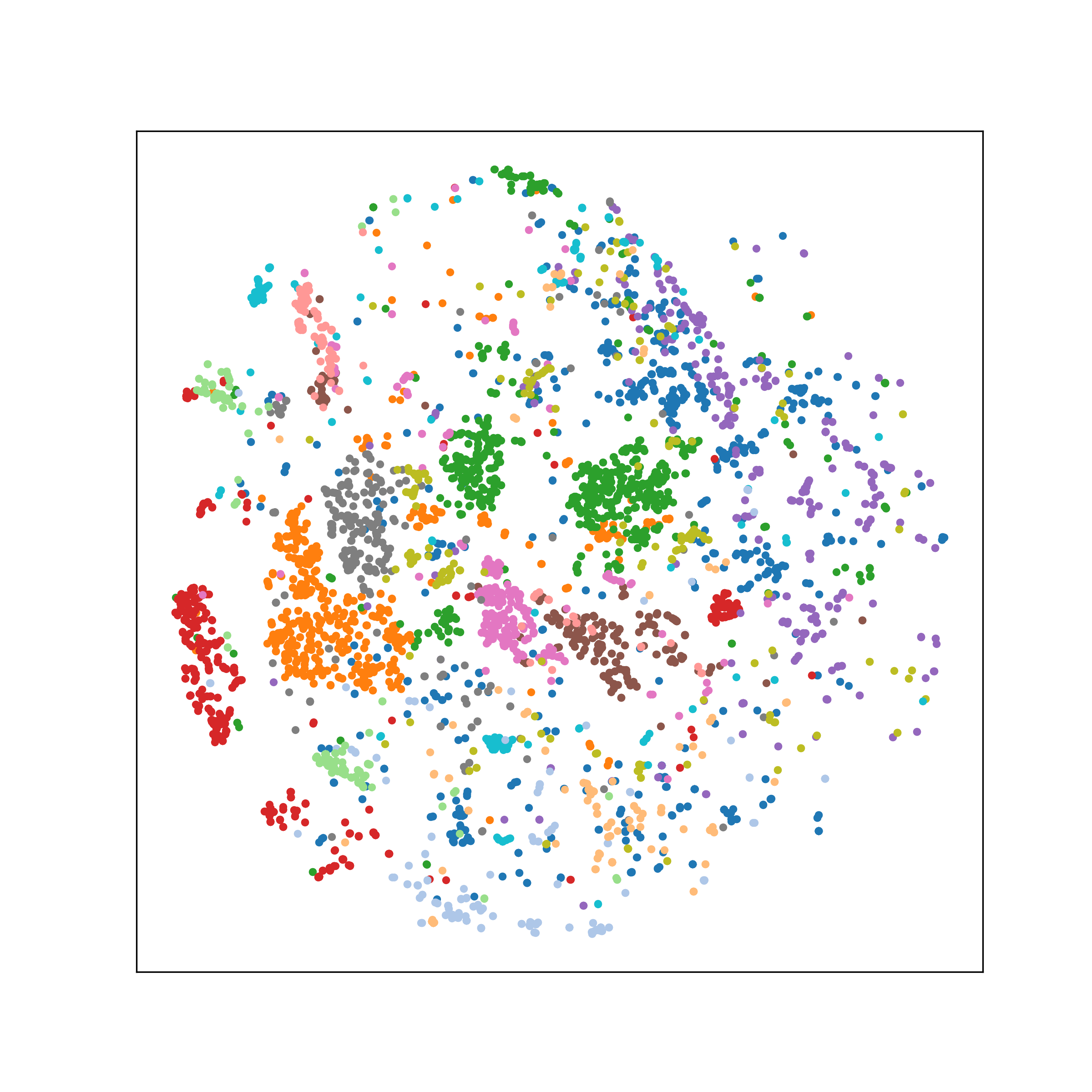}
        \label{fig:lwf}
    }
    \hfill
    \subfloat[PODNet-flat]{
        \centering
        \includegraphics[width=0.31\linewidth]{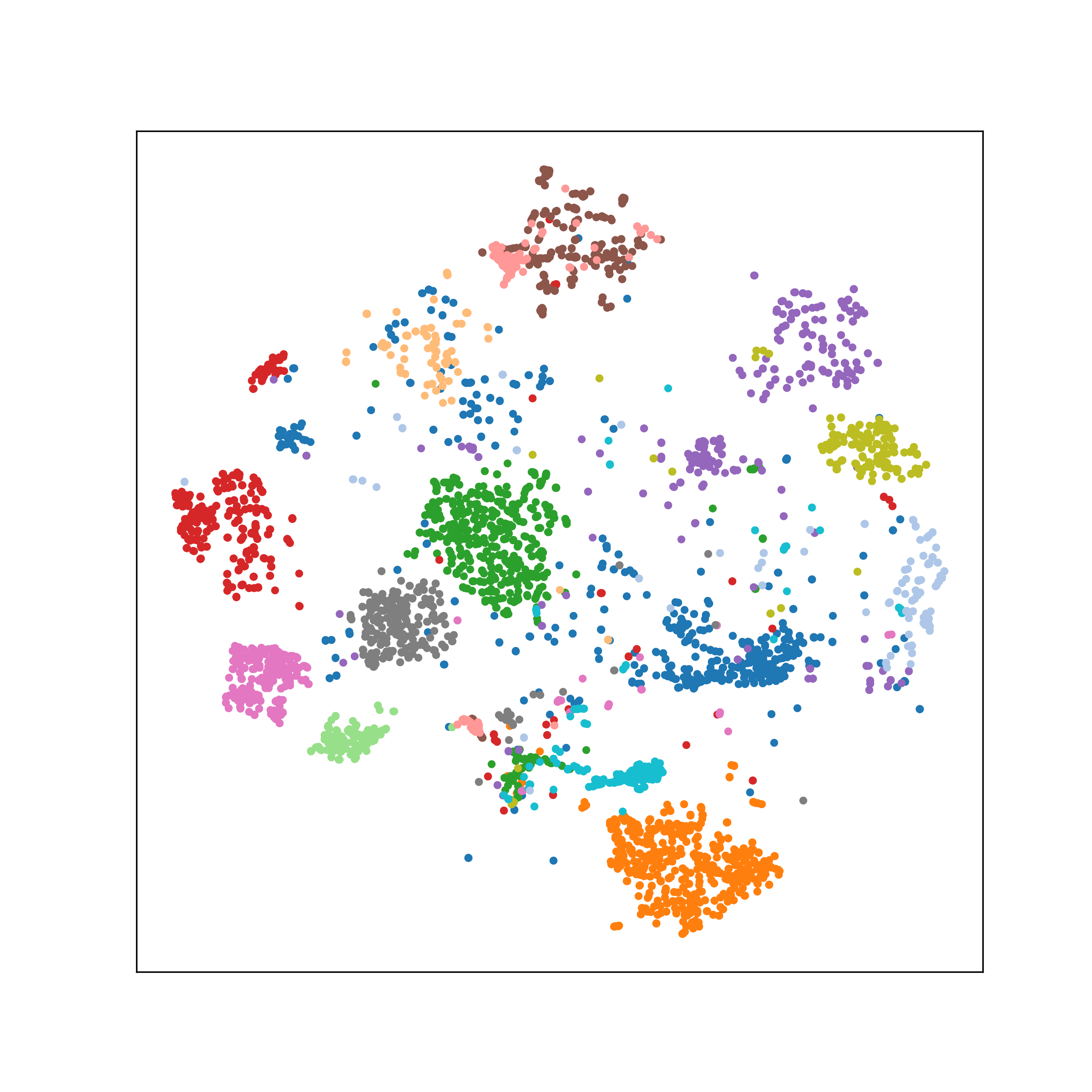}
        \label{fig:podnet}
    }
    \\
    \subfloat[General-Inc]{
        \centering
        \includegraphics[width=0.31\linewidth]{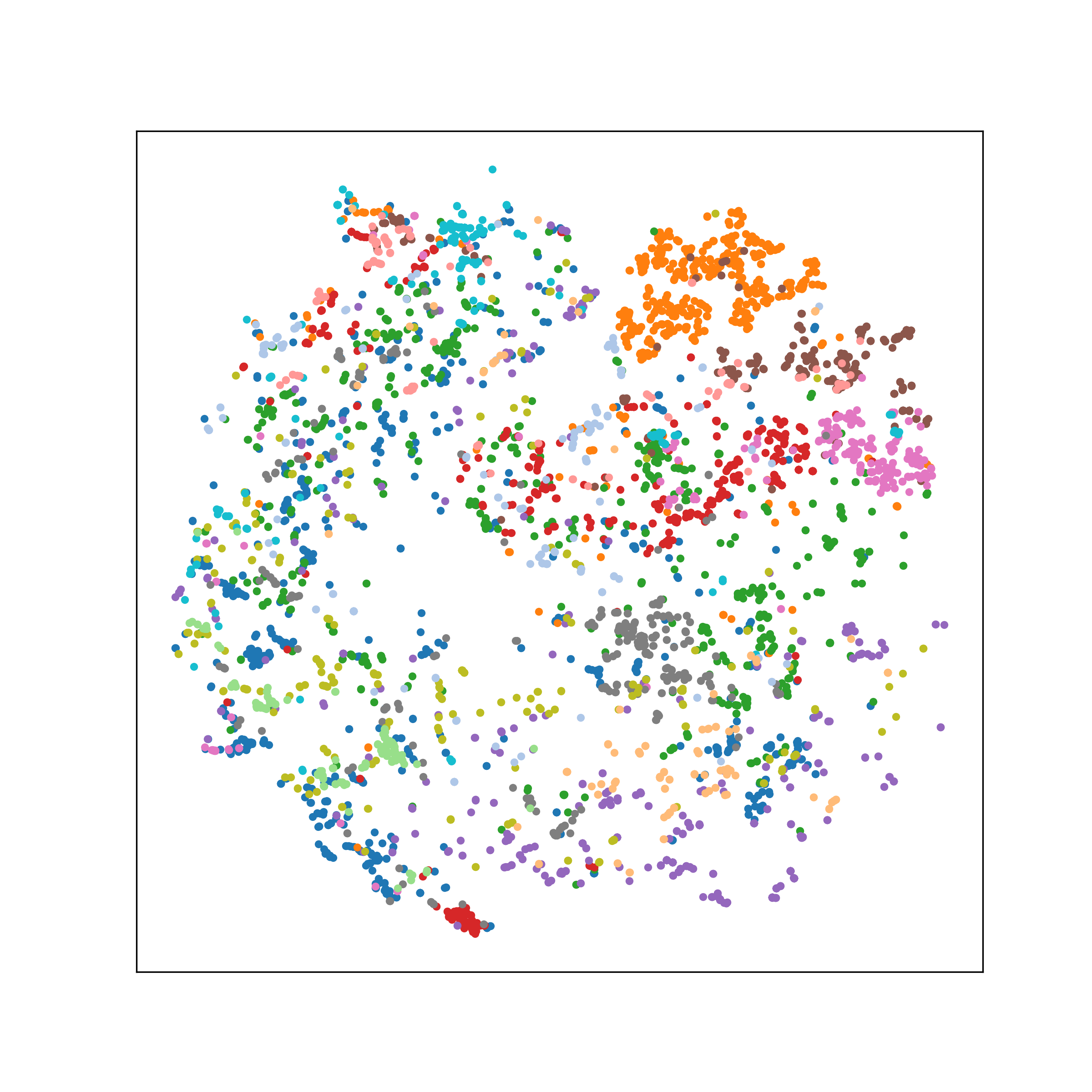}
        \label{fig:gener}
    }
    \hfill
    \subfloat[General-Inc+VCL]{
        \centering
        \includegraphics[width=0.31\linewidth]{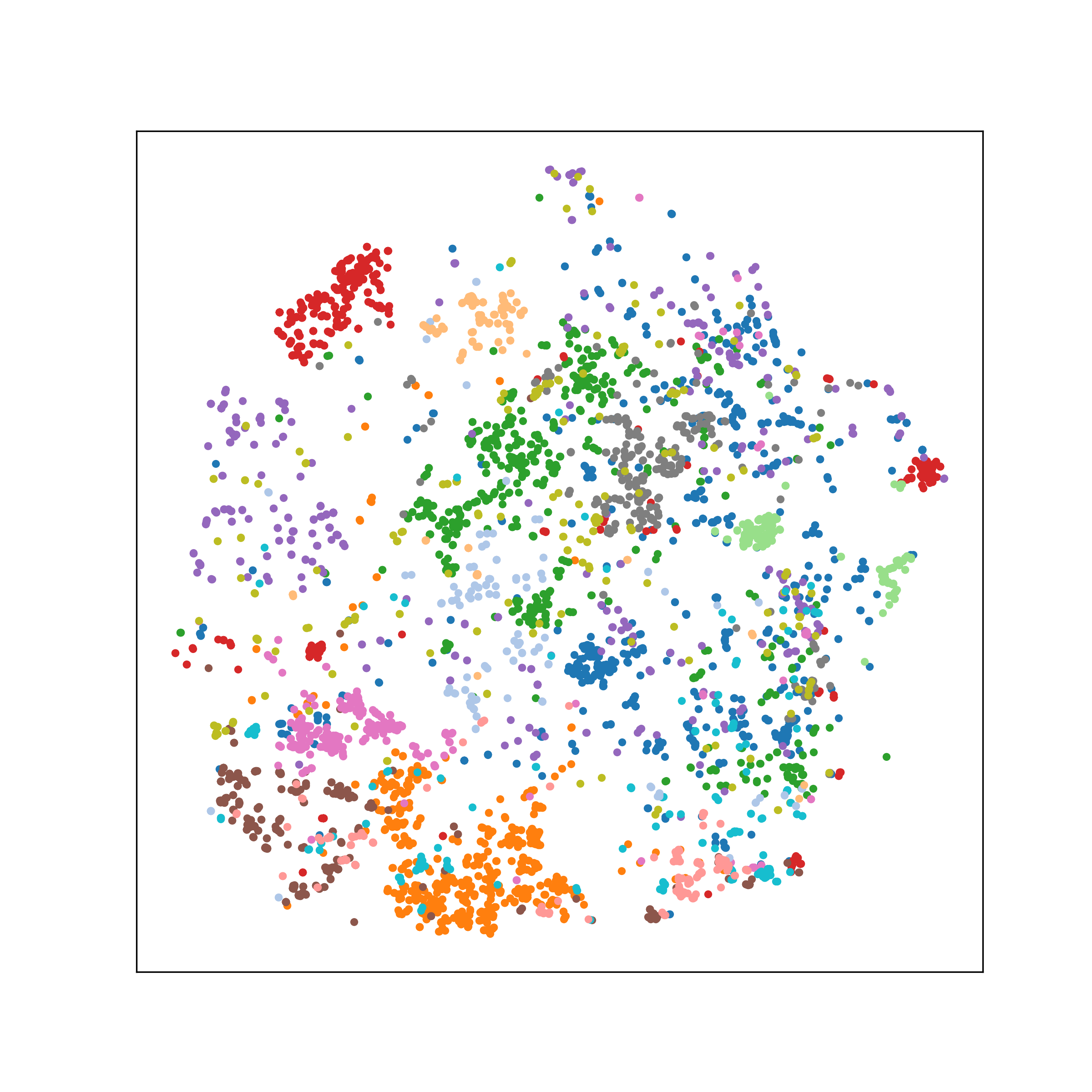}
        \label{fig:gener+vcl}
    }
    \hfill
    \subfloat[General-Inc+SCL]{
        \centering
        \includegraphics[width=0.31\linewidth]{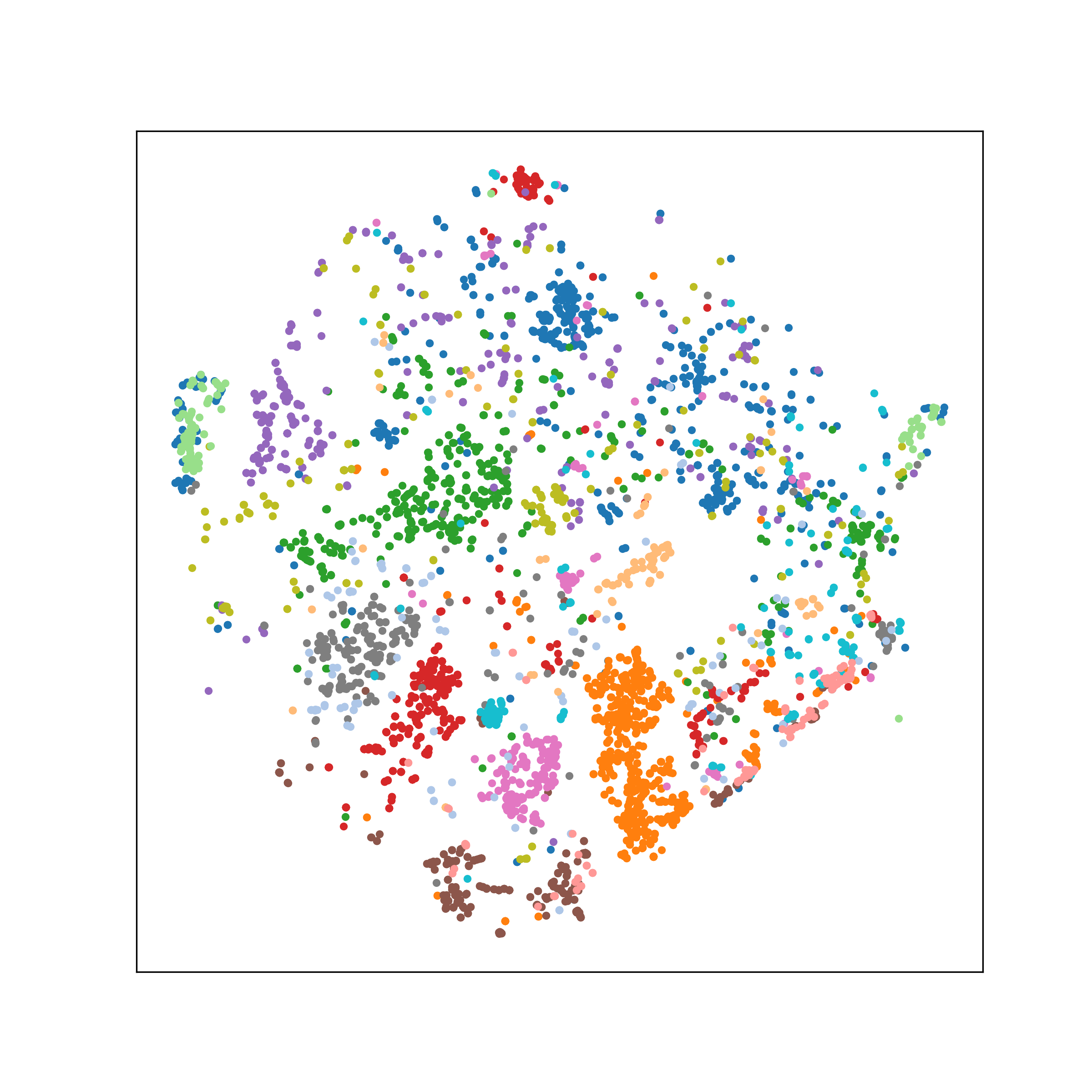}
        \label{fig:gener+scl}
    }
    \\
    \subfloat[PCR]{
        \centering
        \includegraphics[width=0.31\linewidth]{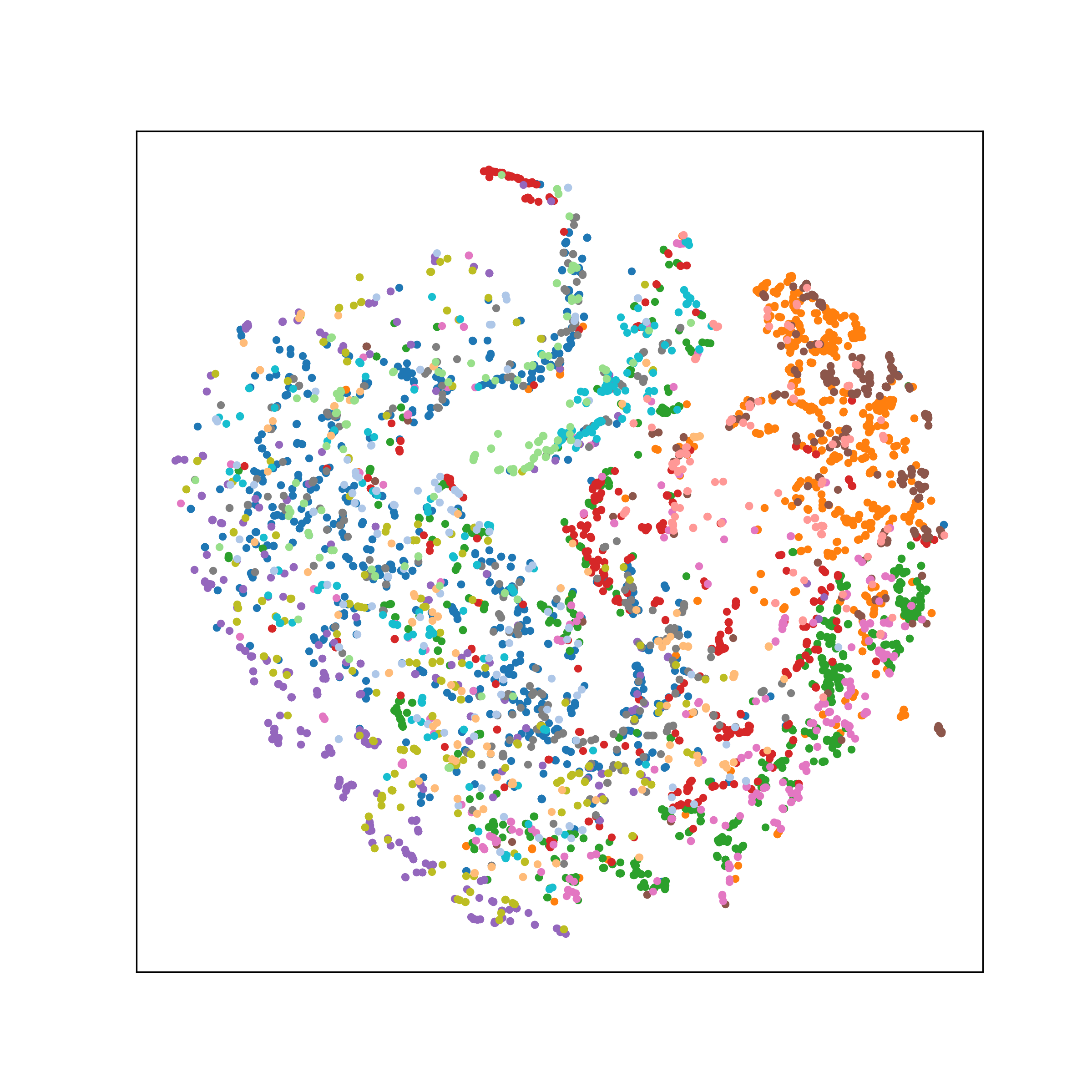}
        \label{fig:pcr}
    }
    \hfill
    \subfloat[PRD]{
        \centering
        \includegraphics[width=0.31\linewidth]{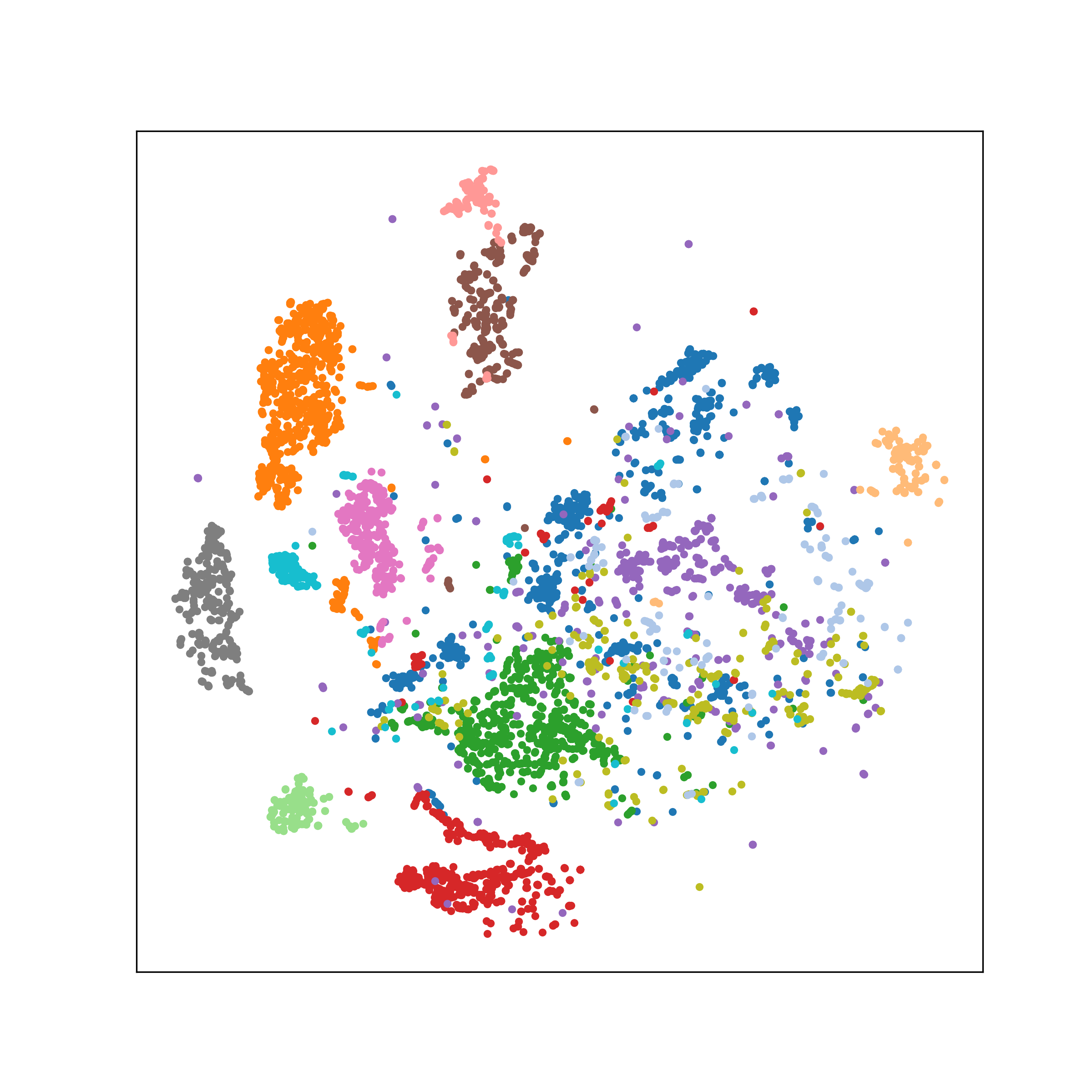}
        \label{fig:prd}
    }
    \hfill
    \subfloat[IRD (Ours)]{
        \centering
        \includegraphics[width=0.31\linewidth]{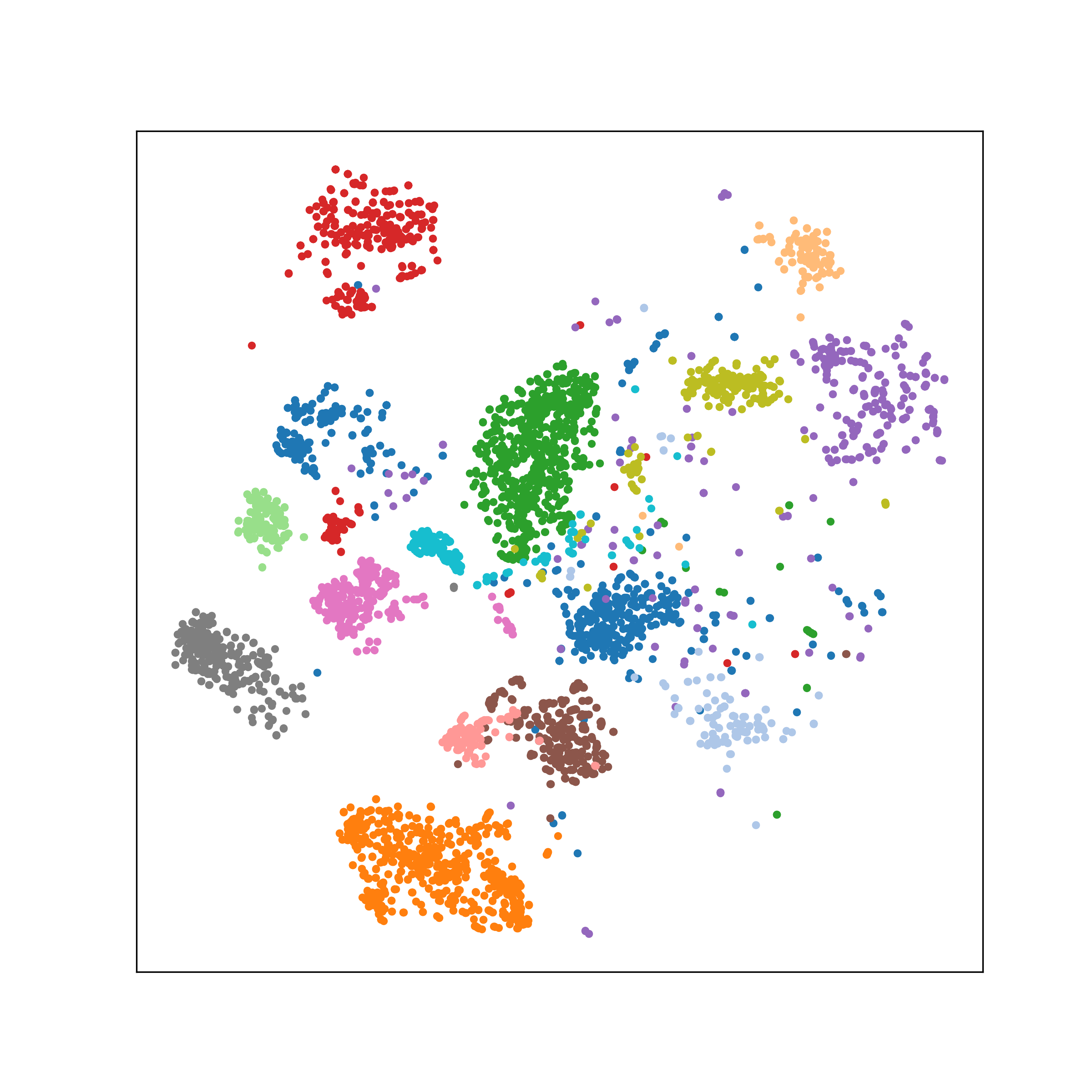}
        \label{fig:ours}
    }
    \caption{t-SNE~\citep{van2008visualizing} visualization on relation features after 5 training phases on HICO-DET.}
    \label{fig:tsne}
\end{figure*}

We utilized the t-SNE visualization technique \citep{van2008visualizing} to demonstrate the robustness and invariance of relation features learned by our method. Fig.~\ref{fig:tsne} shows the t-SNE visualization of relation features from the test set at the final phase, where identical colors indicate features of the same relation category. Our method enables a more compact distribution of features for each relation, suggesting that despite varying HOI classes, the relation features remain consistent across combinations with different objects. This pattern underscores our method's effectiveness in learning relation features invariant to the specific objects involved.

\end{document}